\documentclass[twoside]{article}

\usepackage[preprint]{AISTATS/aistats2026}
\usepackage[round]{natbib}

\usepackage[utf8]{inputenc} 
\usepackage[T1]{fontenc}    
\usepackage{hyperref}       
\usepackage{url}            
\usepackage{booktabs}       
\usepackage{amsfonts}       
\usepackage{nicefrac}       
\usepackage{microtype}      
\usepackage{xcolor}    
\usepackage{amsmath}
\usepackage{bm}

\usepackage{algorithm}
\usepackage{algpseudocode}

\usepackage{caption}
\usepackage{subcaption}

\usepackage{multirow}
\usepackage[table,xcdraw]{xcolor}

\newtheorem{theorem}{Theorem}
\newtheorem{lemma}{Lemma}
\newtheorem{proposition}[theorem]{Proposition}
\newtheorem{corollary}[theorem]{Corollary}

\newtheorem{assumption}{Assumption}
\newtheorem{remark}{Remark}
\usepackage{todonotes}
\usepackage{pifont}
\newcommand*\diff{\mathop{}\!\mathrm{d}}

\def\argmin{\mathop{\arg\min}}

\def\P{\mathbb{P}}
\def\E{\mathbb{E}}
\def\R{\mathbb{R}}

\begin{document}

\twocolumn[

\aistatstitle{Conditional Flow Matching for Bayesian Posterior Inference}

\aistatsauthor{
Percy S.~Zhai$^\ast$ \quad So Won Jeong$^\ast$ \quad  Veronika Ro\v{c}kov\'{a}
}

\aistatsaddress{University of Chicago, Booth School of Business} ]

\renewcommand{\thefootnote}{\fnsymbol{footnote}}
\footnotetext[1]{Equal contribution.}

\begin{abstract}
We propose a generative multivariate posterior sampler via flow matching.
It offers a simple training objective, and does not require access to likelihood evaluation.
The method learns a dynamic, block-triangular velocity field in the joint space of data and parameters, which results in a deterministic transport map from a source distribution to the desired posterior.
The inverse map, named vector rank, is accessible by reversibly integrating the velocity over time.
It is advantageous to leverage the dynamic design: proper constraints on the velocity yield a monotone map, which leads to a conditional Brenier map, enabling a fast and simultaneous generation of Bayesian credible sets whose contours correspond to level sets of Monge-Kantorovich data depth.
Our approach is computationally lighter compared to GAN-based and diffusion-based counterparts, and is capable of capturing complex posterior structures.
Finally, frequentist theoretical guarantee on the consistency of the recovered posterior distribution, and of the corresponding Bayesian credible sets, is provided.
\end{abstract}

\section{INTRODUCTION}

Bayesian inference provides a principled framework of uncertainty quantification in statistical models, enabling estimation of the full posterior distribution.
The primary bottleneck lies in computation, since exact posterior evaluation is often intractable.
Traditional methods rely on approximate inference techniques such as MCMC or variational inference, which may suffer from slow convergence or restrictive assumptions.
For instance, MCMC requires rerunning the entire chain for each new observation.
Recent advances in generative modeling, particularly score-based diffusion models \citep{song2020score, ho2022classifier} and normalizing flows \citep{rezende2016variational}, offer promising alternatives.
However, notable limitations persist: diffusion models require iterative denoising steps and rely on approximating the score function, while normalizing flows impose invertibility constraints and exact likelihood evaluations.

This work develops a novel Bayesian posterior sampler based on \emph{flow matching} \citep{lipman2022flow, liu2022flow}, a generative modeling technique that learns a deterministic velocity field transporting a simple source distribution to a complex target.
Consider parameter space $\Theta \subset \R^d$ and data space $\mathcal Y \subset \R^n$.
Observations $y \in \mathcal Y$ arise from parameters $\theta \in \Theta$, either via an explicit likelihood $L(y \mid \theta)$ or through an implicit simulator.
We allow entries of $y$ to be potentially dependent.
Combining the data-generating process with a prior simulator $\pi(\theta)$, one can generate joint samples $\{y_i, \theta_i\}_{i=1}^N$, where the size of the synthetic joint pairs $N$ can be large at a low computational cost.
This setup enables conditional inference in a likelihood-free setting, also known as simulation-based inference (SBI).

For a fixed observation $y^* \in \mathcal Y$, the primary objective of Bayesian inference is to sample from the posterior $\pi(\theta \mid y^*)$.
From the optimal transport point of view, this is equivalent to learning a transport map from a source distribution (e.g. uniform) to the target posterior.
This perspective has been considered by multiple works in the literature.
\cite{wang2023adversarialbayesiansimulation} learns the posterior sampler by generative adversarial networks (GANs).
A similar route is taken by \cite{baptista2024conditional} using the Monotone GAN (M-GAN) approach.
More recently, \cite{kim2024deep} develops a conditional quantile learning method that recovers an optimal transport map on the parameter space $\Theta$.
Despite the considerably different approaches, the latter two both yield the conditional Brenier map \citep{carlier2017vector} under monotonicity assumptions. 
This paper marks an addition to the toolbox for posterior sampling from another perspective, using the flow matching technique to achieve a dynamic extension. 
See Table~\ref{tab:model_comparison} for a comparison of selected approaches, and Appendix~\ref{app:related_works} for a comprehensive literature review.

\begin{table*}
\centering
\newcommand{\greencheck}{{\color{green!60!black}\checkmark}}
\newcommand{\redx}{{\color{red}\ding{55}}}
\newcommand{\orangetilde}{{\color{orange!90!black}\Large$\sim$}}
\resizebox{\textwidth}{!}{
\begin{tabular}{@{}l c c c c@{}}
\toprule
 &
\textbf{Flow Matching} &
\textbf{GANs} &
\textbf{Diffusion Models} &
\textbf{Quantile NN} \\
\textbf{Feature} &
\textbf{(Ours)} &
\textbf{(e.g., \citet{baptista2024conditional})} &
\textbf{(e.g., \citet{chung2022diffusion})} &
\textbf{(e.g., \citet{kim2024deep})} \\
\midrule

\textbf{Core Principle} & 
\begin{tabular}[t]{@{}l@{}}Learns a velocity field (ODE) to \\ transport noise to the joint distribution.\end{tabular} & 
\begin{tabular}[t]{@{}l@{}}Learns a generator to fool a \\ discriminator on joint samples.\end{tabular} & 
\begin{tabular}[t]{@{}l@{}}Learns to reverse a forward \\ noising process via score estimation.\end{tabular} & 
\begin{tabular}[t]{@{}l@{}}Directly learns the conditional \\ quantile function (Brenier map).\end{tabular} \\
\addlinespace

\textbf{Training Objective} & 
Regression Loss. &
Adversarial Loss. & 
Score-Matching Loss.& 
Pinball Loss or MK objective\\

\textbf{Likelihood-Free} & \greencheck & \greencheck & \orangetilde & \greencheck \\
\addlinespace
\textbf{Access to the Inverse Map} & \greencheck & \redx & \orangetilde & \redx \\
\addlinespace
\textbf{Uncertainty Quantification} & \greencheck & \orangetilde & \redx & \greencheck \\

\bottomrule
\multicolumn{5}{l}{\greencheck: Supported Directly \quad \orangetilde: Possible with Constraints/Indirectly \quad \redx: Not Directly Supported} \\
\end{tabular}
}
\caption{Comparison of selected generative approaches for posterior sampling.}
\label{tab:model_comparison}
\end{table*}

Our approach learns the posterior by matching the \emph{joint distribution} of parameters and observations, rather than the marginal or conditional alone \citep{mirza2014conditional, zhou2023deep}.
The theoretical cornerstone is the \emph{block-triangular map}.
We consider maps $T: \mathcal Y \times \Theta \rightarrow \mathcal Y \times \Theta$ that jointly transports random noise $y_0\in\mathcal Y$ and $\theta_0\in\Theta$ to the data denoted by $y_1 \in \mathcal Y$ and the parameter $\theta_1 \in \Theta$ at the following form,
\begin{equation}\label{eq:triangular.map}
    T(y_0,\theta_0) = (F(y_0), G(F(y_0),\theta_0)),
\end{equation}
where $F: \mathcal Y \rightarrow \mathcal Y$, and $G: \mathcal Y \times \Theta \rightarrow \Theta$.
Such maps are called \emph{block-triangular}, in a sense that the Jacobian matrix of the joint transport map has a lower block-triangular shape.
A primary theoretical result established in \cite{baptista2024conditional} states that if a joint source distribution $\eta$ is transported by such a map $T$ to the joint distribution (i.e. the push-forward measure $T_\sharp \eta = \pi(y,\theta)$), then $G(y^*,\cdot)$ maps the $\Theta$-marginal of the source distribution to the posterior distribution of $\theta$.
Due to this convenient result, the task of posterior sampling boils down to learning a joint map $T$ that has a block-triangular structure.
The M-GAN developed by \cite{baptista2024conditional} learns the parameter map $G(y^*, \cdot)$ through an adversarial scheme.

In this paper, we instead learn the block-triangular map $T$ using the flow matching method.
That is, for any $t\in[0,1]$, we learn a joint velocity field in $\mathcal Y \times \Theta$ that results in a block-triangular transport map from $t=0$ to $t=1$.
The dynamic nature of this method leads to a convenient training process.
For example, we may choose any interpolation path to reduce computational complexity while obtaining a theoretical guarantee.
When the map is monotone, both \cite{baptista2024conditional} and \cite{kim2024deep}, as well as our method, eventually learn the conditional Brenier map, an optimal transport map from a source distribution to the posterior distribution.
With this same goal in mind, \cite{baptista2024conditional} leverages adversarial learning scheme through GAN, and \cite{kim2024deep} relies on a variation of Quantile Neural Network (QNN), while our method leverages flow matching architecture to obtain the conditional map.
See Table~\ref{tab:model_comparison}
for a detailed comparison.
Recognizing the importance of the block-triangular maps,
a followup work of the M-GAN method \citep{alfonso2023generative} formulated the problem of learning such a map by a discretized temporal mapping method.
We highlight that our route reaching such maps is distinctive of previous works in that our methods rely on flow matching unlike GANs, and operate on continuous time space unlike \cite{alfonso2023generative}.

After finalizing our manuscript, we became aware of concurrent works by \citet{kerrigan2024dynamic} and \citet{chemseddine2024conditional}, 
which learn a conditional velocity field with the observation treated as a fixed input, aiming for optimal transport optimality.
Our work, by contrast, learns a block-triangular velocity field on the joint space.
Moreover, our proposed method aims for consistency of posterior distributional estimation, as opposed to optimal transport from source to target.
A detailed comparison is provided in Appendix~\ref{sec:comparison.to.concurrent}.

\paragraph{Main Contributions}
\begin{enumerate}

    \item \textbf{Likelihood-free Posterior Sampler.} In Section~\ref{sec:methodology}, we adapt flow matching technique to the posterior sampling problem, with or without access to the explicit likelihood.
    With a similar accuracy in posterior sampling, our proposed method requires less computation time than MCMC and M-GAN.
    Our proposed method is a dynamic extension of block-triangular maps, with a straightforward access to the inverse maps.
    
    \item \textbf{Inference through Credible Sets.} 
    Our posterior sampling method enables us to implement posterior uncertainty quantification (Section~\ref{sec:monotonicity}).
    Under mild assumptions,
    we can draw Bayesian credible sets at multiple levels simultaneously without having to learn the maps repeatedly.
    The contour of these credible sets agree with the levels of Monge-Kantorovich data depth, providing a more sensible shape.
    Our method also allows an easy access to the inverse map, enabling a conditional rank function that plays a role similar to the p-value in comparing parameters to the posterior distribution.

    \item \textbf{Consistency of Learned Posterior.}
    In Section~\ref{sec:theory}, we provide theoretical guarantee on the asymptotic consistency of distributions recovered by flow matching if the velocity is learned by multilayer perceptron with ReLU activation functions, the first of its kind to the best of our knowledge.
    We establish that the recovered posterior distribution converges to the true posterior in 2-Wasserstein distance.
    As a corollary, the corresponding Bayesian credible sets are also consistent in Hausdorff distance.
\end{enumerate}

\section{METHODOLOGY}\label{sec:methodology}

We develop a method based on the theory on block-triangular maps \eqref{eq:triangular.map},
but substantially different from M-GAN \citep{baptista2024conditional}.
First, we extend the block-triangular map to a dynamic domain.
Instead of learning a one-off transport map, our proposed method learns a dynamic map over time $t\in[0,1]$.
We then have a freedom on choosing the intermediate states (i.e., interpolation path).
Second, unlike several existing posterior sampling methods that learns the posterior transport map directly, our proposed method is based on learning a transport map on the \emph{joint space} $\mathcal Y \times \Theta$.
This decoupling of marginal transport from conditional transport brings extra flexibility by alleviating representational burden on the neural network for learning the velocity field.
A more detailed discussion in this regard is deferred to Appendix \ref{sec:comparison.to.concurrent}.
Our proposed method also comes with several byproducts, including the access to the data map $F$ and the inverse maps $F^{-1}$ and $[G(y^*,\cdot)]^{-1}$.

To recover the posterior distribution, it is only required that the final trained map $T$ is block-triangular as in \eqref{eq:triangular.map}.
By Theorem 3.4 in \cite{baptista2024conditional}, the map $G(y^*,\cdot)$ transports the noise $\theta_0$ to the posterior distribution $\theta_1 \sim \pi(\cdot \mid y^*)$.
We aim to learn the joint map $T$ via flow matching, which is characterized by a dynamic velocity field for $t\in[0,1]$ (the ending time is set to $1$ hereafter, without loss of generality).
The joint transport map from the source $(y_0,\theta_0)$ to the target $(y_1,\theta_1)$ is recovered by accumulating the velocity over time.

\subsection{Learning the Joint via Flow Matching}\label{sec:joint.FM}

For brevity, denote $x_t = (y_t, \theta_t)$ for any time point $t\in [0,1]$.
Our goal is to learn a transport map from the source distribution $x_0\sim p_0$ to the target joint distribution $x_1 \sim p_1 = \pi(y_1,\theta_1)$.
This can be done by Flow Matching (FM, \cite{lipman2022flow, liu2022flow, tong2023improving}), which learns a deterministic velocity field that governs the time-dependent flow of $x_t$.
In general, we denote a \emph{flow} starting in time $t_0$ by $\{x_{t_0,t}\}_{t\in[t_0,1]}$.
In our problem, $t_0=0$, with the initial point $x_{0,0} = x_0$, and $x_{0,t} = x_t$.
For simplicity, we shall denote the flow by $\{x_t\}_{t\in[0,1]}$ hereafter, unless otherwise specified.
Each flow can be identified with a \emph{velocity field}, which is the derivative of $x_t$ with respect to time, $v_t(x_t) = \diff x_t/\diff t$.
Each intermediate state, $x_{t}$, can thus be regarded as a collective result of the velocity $\{v_s\}_{s\in[0,t]}$, which also defines a dynamic push-forward map $T_t$ applied from the origin $x_0$.
In this case, $T_1$ corresponds to the final resulting map $T$ in \eqref{eq:triangular.map}.

Denote the velocity field $\{v_t\}_{t\in[0,1]}$ as a function $v: \mathcal X\times [0,1] \rightarrow \R^{n+d}$.
The FM loss function is defined as
\begin{equation}\label{eq:flow-matching-loss-intractable}
\mathcal{L}_{\text{FM}}(\hat v, v) = \int_0^1 \mathbb{E}_{x \sim p_t} \| \hat{v}_t(x) - v_t(x) \|^2 \diff t.
\end{equation}
However, this objective is generally intractable as we do not have an access to the unknown true velocity $v_t$.
To make this objective more feasible, \cite{lipman2022flow} proposes a \emph{conditional} flow matching loss.
The idea is to construct $p_t$ implicitly as a mixture of conditionals
\begin{equation*}
  p_t(x) = \int p_t(x \mid z) q(z) \diff z,  
\end{equation*}
where $z$ is an auxiliary conditioning variable (often taken as $ z = (x_0, x_1)$) and $q(z)$ is a known joint distribution over pairs. A standard instantiation is the Gaussian interpolation path 
\begin{equation}\label{eq:gaussian-path}
 x_t = \mu_t(z) + \epsilon, \quad \epsilon \sim \mathcal{N}(0, \sigma_t^2 I),
\end{equation}
where $\mu_t(z) = (1 - t)x_0 + t x_1 $ defines a linear interpolation, and $\sigma_t > 0$ controls the spread of the conditional. 
The velocity field conditioned on the initial and ending states can be written explicitly as 
\begin{equation*}
    v_t(x\mid x_0, x_1) =\frac{\sigma_t'}{\sigma_t} (x - \mu_t) + \mu_t',
\end{equation*}
where $\mu_t'$ and $\sigma_t'$ denote the time derivative of $\mu_t$ and $\sigma_t$, respectively.
This yields a tractable target for regression under the conditional model.
In practice, we use the deterministic limit $\sigma_t \to 0$, yielding the straight-line path $x_t=(1-t)x_0+tx_1$ with conditional velocity $v_t(x \mid x_0, x_1) = x_1 - x_0$.

The conditional flow matching loss is then written as
\begin{equation}\label{eq:cfm-loss}
 \mathcal{L}_{\text{CFM}}(\hat v, v) = \int_0^1 \mathbb{E}_{q(z), p_t(x|z)} \| \hat{v}_t(x) - v_t(x|z)  \|^2 \diff t,
\end{equation}
which is fully computable given samples from \( q(z) \) and the interpolation path. 
\cite{lipman2022flow} showed that (\ref{eq:cfm-loss}) is equivalent to the unconditional flow matching loss \eqref{eq:flow-matching-loss-intractable} up to a constant.
In other words, the gradient with respect to neural network parameters $\varphi$ for both loss functions \eqref{eq:flow-matching-loss-intractable} and \eqref{eq:cfm-loss} matches:
\begin{equation}\label{eq:equivalence-fm-cfm}
    \nabla_\varphi \mathcal{L}_\text{FM} = \nabla_\varphi \mathcal{L}_\text{CFM}.
\end{equation}
By learning the minimizer of $\mathcal L_{\text{CFM}}$, we obtain the velocity field $\hat v_t(x)$ that also minimizes $\mathcal L_{\text{FM}}$.

\subsection{Dynamic Block-triangular Map}

To apply the block-triangular structure on the dynamic joint maps learned by FM,
an analytically and practically convenient setup is to constrain $T_t$ to be block-triangular for all $t\in[0,1]$,
\begin{equation}\label{eq:dynamic.triangular.map}
    T_t(y_0,\theta_0) = (F_t(y_0), G_t(F_t(y_0), \theta_0)),
\end{equation}
where for any $t\in[0,1]$, $F_t: \mathcal Y \rightarrow \mathcal Y$ maps $y_0$ to $y_t$, and $G_t: \mathcal Y \times \Theta \rightarrow \Theta$ maps the initial pair $x_0=(y_0,\theta_0)$ to $\theta_t$.
Under these notations, the final maps $F_1$ and $G_1$ correspond to $F$ and $G$ respectively.

In fact, the maps $T_t$ in \eqref{eq:dynamic.triangular.map} can be achieved by accumulating block-triangular velocity of the form
\begin{equation}\label{eq:triangular.velocity}
    \frac{\diff y_t}{\diff t} = f_t(y_t), \quad \frac{\diff \theta_t}{\diff t} = g_t(y_t, \theta_t)
\end{equation}
from time zero to time $t$.
The correspondence between the univariate ODE $\diff y_t / \diff t = f_t(y_t)$ and its solution $y_t = F_t(y_0)$ is straightforward.
It is therefore possible to write any $y_t$ as a function of $y_1$:
instead of pushing forward from $t=0$ to $t=1$, we can reversely push from $t=1$ back to $t=0$.
Specifically, define
\begin{equation}\label{eq:ft.tilde.def}
\tilde F_t(y_1) = y_t = F_t(y_0) = F_t(F^{-1}(y_1)).
\end{equation}
Then the second half of \eqref{eq:triangular.velocity} can be written as $\frac{\diff \theta_t}{\diff t} = g_t(\tilde F_t(y_1), \theta_t)$, yielding a univariate ODE for $\theta_t$.
Here, $\theta_0$ is a fixed value realization of a stochastic noise in the parameter space, and $y_1$ is treated as a constant.
The solution of this ODE can be written as $\theta_t = G_t(y_1,\theta_0)$, which is equivalent to the form in \eqref{eq:dynamic.triangular.map}.
This essentially proves the following important lemma.
\begin{lemma}\label{lm:map.velo}
    The dynamic map $T_t$ is block-triangular in the form of \eqref{eq:dynamic.triangular.map} for all $t\in [0,1]$ if the velocity field can be expressed in the block-triangular form of \eqref{eq:triangular.velocity}.
\end{lemma}
This means that a block-triangular velocity design will always lead to block-triangular dynamic maps throughout the process, including the resulting map at $t=1$.
Based on \eqref{eq:triangular.velocity} and \eqref{eq:ft.tilde.def}, it is therefore obvious that given a fixed $y^* \in \mathcal Y$, the posterior can be simulated by pushing a random noise $\theta_0$ through the learned velocity field,
\begin{equation}\label{eq:conditional.integral}
    G(y^*, \theta_0) = \theta_0 + \int_0^1 g_t(\tilde F_t(y^*), \theta_t) \diff t.
\end{equation}
If we obtain a velocity field accurately enough, by sampling several realizations of $\theta_0$ from the source distribution and passing them through the map $G(y^*,\cdot)$, the resulting data points $\theta_1$ should follow the posterior distribution $\pi(\cdot \mid y^*)$.
\eqref{eq:conditional.integral} plays a central role in our proposed algorithm for posterior simulation.

A major advantage of this dynamic configuration is the automatic invertibility of the learned map.
With a fixed observation $y^*$, the inverse of the conditional map $G(y^*,\cdot)$ can also be obtained by integrating \eqref{eq:conditional.integral} in reverse.
Without additional monotonicity assumptions, these inverse maps are more intuitive than useful.
We defer further discussions on inverse maps to Section \ref{sec:monotonicity}.
Our entire posterior learning framework is summarized in Algorithm~\ref{alg:triangular-flow-posterior}.

\begin{remark}
    From the first sight, it may seem that the class of admissible joint flows is reduced.
    Indeed, the block-triangular map \eqref{eq:dynamic.triangular.map} is a smaller class compared to a fully general joint map.
    This restriction, however, does not affect the representation capability for the final joint distribution $p_1$.
    Any target joint law of $(\theta, y)$ can be represented by transporting through the marginal of $y$ and then through the conditional law $\theta \mid y$.
    Rather than on the final joint distribution, the major restriction is essentially on the choice of intermediate flow path for $t\in(0,1)$, rather than the ability to represent the target distribution itself.
\end{remark}

\begin{algorithm}
\caption{Posterior Sampling with Flow Matching}
\label{alg:triangular-flow-posterior}
\begin{algorithmic}[1]
\State {\bf Input:} Joint samples $\{x_1^{(i)}\}_{i=1}^N$ where ${x_1 = (y_1, \theta_1)}$, from the target distribution $p_1(x)$, a source distribution $p_0(x)$, and a neural velocity field $u_\varphi(x, t)$.
\State {\bf Parameters:} Training steps $S$, learning rate $\eta$, batch size $B$
\vspace{0.5em}
\For{$s = 1, \dots, S$}
    \State Sample minibatch $\{x_1^{(i)}\}_{i=1}^B \sim p_1(x)$
     \State Sample source points $\{x_0^{(i)}\}_{i=1}^B \sim p_0(x)$, and a random time $t^{(i)} \sim \mathrm{Unif}[0,1]$ for each instance $i$.
    \State Form interpolation: $x_t^{(i)} = (1 - t^{(i)}) x_0^{(i)} + t^{(i)} x_1^{(i)}$
    \State Compute target conditional velocity:
    \begin{equation*}
        v^{(i)} = x_1^{(i)} - x_0^{(i)}.
    \end{equation*}
    \State Predict velocity: $\hat v^{(i)} = u_\varphi \big(  x_t^{(i)},  t^{(i)} \big)$
    \State Compute loss: $\mathcal{L} = \frac{1}{B}\sum_{i=1}^B \| \hat v^{(i)} - v^{(i)} \|^2$
    \State Update parameters: $\varphi \leftarrow \varphi - \eta \nabla_\varphi \mathcal{L}$
\EndFor
\vspace{0.5em}
\State {\bf Sampling posterior given $y^\ast$:}
\State Draw $x_0 = (y_0, \theta_0)\sim p_0(x)$
\State Solve (\ref{eq:conditional.integral}) or an equivalent ODE: $\frac{\diff \theta_t} {\diff t} = u_\varphi (y^*, \theta_0, t)$ for $t \in [0,1]$ to obtain $\hat G_t(y^*, \cdot)$
\State Return $\theta_1$ as posterior sample
\end{algorithmic}
\end{algorithm}

\begin{remark}
In Algorithm~\ref{alg:triangular-flow-posterior}, line 6 employs linear interpolation for simplicity, but our framework is agnostic to the choice of conditional path. 
Any valid conditional path can be used, including the optimal transport path \citep{tong2023improving} or the widely used Gaussian path \eqref{eq:gaussian-path}.
\end{remark}

\subsection{Monotonicity, Monge-Kantorovich Depth, and Inverse Maps}\label{sec:monotonicity}
We can already construct a posterior sampler via the block-triangular map that recovers the true underlying posterior, $\pi(\cdot \mid y^*)$.
From an optimal transport perspective, there exists multiple maps that push a source distribution on $\Theta$ to the true posterior.
This lack of uniqueness does not affect the accuracy of posterior sampling.
However, if we intend to make use of the recovered posterior distribution for inference tasks (for example, to create Bayesian credible sets with meaningful interpretation), then it is convenient to assume that the transport map $G(y^*,\cdot)$ is monotone, i.e. for any $\theta_0, \theta_0' \in \Theta$,
\begin{equation*}
    (G(y^*, \theta_0) - G(y^*,\theta_0') )^\top (\theta_0-\theta_0') \geq 0.
\end{equation*}
See, e.g. \cite{chernozhukov2017monge}.
We shall see that such monotonicity enables nested credible sets at a low computational cost.
One method to enforce this in the flow matching setup is to constrain the $\Theta$-coordinate of the velocity, $g_t(y_t,\theta_t)$, to the monotone functions in $\theta_t$.
One rigorous method to achieve this is to let
\begin{equation}\label{eq:ICNN.representation}
    g_t(y,\theta) = \nabla_\theta \psi_t(y, \theta),
\end{equation}
where $\psi_t(y,\theta)$ is convex in $\theta$, and train the convex function $\psi_t$, for instance, with input convex neural network (ICNN, \cite{amos2017inputconvexneuralnetworks}).
This implementation represents a key aspect of our approach that is not explored in the concurrent work \citep{kerrigan2024dynamic}.

Under the monotonicity assumption, let us specifically suppose that the source distribution of $\theta_0$ is spherical uniform, i.e. $\theta_0 = r\phi$, with a unidimensional $r\sim \text{Unif}(0,1)$, and a $(d-1)$-dimensional $\phi$ following uniform distribution on the unit sphere $\mathcal S^{d-1}(1)$.
Then the map $G(y^*,\cdot)$ can be regarded as a multivariate Monge-Kantorovich (MK) conditional vector quantile, also named Brenier map, denoted by $Q_{\theta\mid {y^*}}: \Theta\rightarrow\Theta$.
In this case, the Bayesian $\tau$-credible set can be established by passing a ball $S^d(\tau)$ through the learned transport map obtained from Algorithm \ref{alg:triangular-flow-posterior},
\begin{equation}\label{eq:credible.set}
\hat C_\tau(y^*) = \hat G(y^*, S^d(\tau)).
\end{equation}
We can obtain multiple nested credible sets by passing balls of different radii through $\hat Q_{\theta\mid {y^*}}$, without resampling or relearning the map, significantly speeding up the computation process.
We shall see that under some regularity conditions, these credible sets $\hat C_\tau(y^*)$ are consistent with the oracle sets,
\begin{equation}\label{eq:oracle.set}
    C_\tau(y^*) = G(y^*, S^d(\tau)).
\end{equation}
Interestingly, these oracle sets are meaningful in that each $C_\tau(y^*)$ is identical to the Monge-Kantorovich depth region with probability content $\tau$ \citep{chernozhukov2017monge}, which is a multivariate analogue of quantiles.

Since this map is learned via flow matching, we automatically have access to the inverse map of $Q_{\theta\mid y^\ast}$, called Monge-Kantorovich conditional vector rank \citep{chernozhukov2017monge}, denoted by
\begin{equation}\label{eq:MK.rank}
R_{\theta\mid y^*}(\theta_1) = [G(y^*, \cdot)]^{-1}(\theta_1).
\end{equation}
The MK conditional vector rank can be decomposed into a MK conditional rank function $r_{\theta\mid y^*}: \Theta\rightarrow [0,1]$, with $r_{\theta\mid y^*}(\theta_1) = \|R_{\theta\mid y^*}(\theta_1)\|$, and a MK conditional sign function $u_{\theta\mid y^*}$, mapping any parameter $\theta_1 \in \Theta$ to $u_{\theta\mid y^*}(\theta_1) = R_{\theta\mid y^*}(\theta_1) / r_{\theta\mid y^*}(\theta_1) \in \mathcal S^{d-1}(1)$.
Note that $r_{\theta\mid y^*}$ is a representation of the Monge-Kantorovich depth of the parameter in the posterior distribution $\pi(\cdot \mid y^*)$.
For any $\theta_1, \theta_1' \in \Theta$, the MK depth of $\theta_1$ is greater or equal to that of $\theta_1'$ if and only if $r_{\theta\mid y^*}(\theta_1) \leq r_{\theta\mid y^*}(\theta_1')$.
Moreover, $r_{\theta\mid y^*}$ plays a role similar to the p-value -- when this quantity is close to one, it is unlikely that the corresponding $\theta$ is sampled from the posterior distribution.

\begin{remark}\label{rmk:ICNN.representation}
    The representation \eqref{eq:ICNN.representation} using ICNN comes with a caveat: the class of transport maps $G$ representable by such a velocity $g_t$ is in fact restricted, in that only expansive maps ($\nabla_\theta G \succeq I$) can be represented.
    This limitation can be practically mitigated by initializing the source distribution with a much smaller variance than the target.
    We provide reasoning of this claim and an alternative velocity formulation in Appendix \ref{sec:ICNN.representation.issue}.
\end{remark}

\section{THEORETICAL GUARANTEE}\label{sec:theory}
This section is aimed at establishing consistency results on the estimated posterior.
By mapping the source distribution through the learned map $\hat G(y^*,\cdot)$, we obtain an estimated posterior $\hat\pi(\theta\mid y^\ast)$.
We start by showing that it is close to the true posterior $\pi(\theta\mid y^\ast)$.
Under regularity conditions, we establish an asymptotic consistency result on the posterior.
With extra monotonicity assumptions, we establish the consistency of Bayesian credible sets based on the theoretical framework of Monge-Kantorovich quantiles. We defer the proofs of the results presented in this section to Appendix~\ref{app:proof}.

\subsection{Block Triangular Mapping Guarantee}
We start by imposing the following mild technical assumptions on the true underlying velocity field, $v_t$, and its estimate by the algorithm, $\hat v_t$.
\begin{assumption}[Smoothness of Flows]\label{ass:flow.smoothness}
    For each $\xi\in\mathcal Y\times \Theta$ and $t_0\in[0,1]$ there exist unique flows $\{\hat x_{t_0,t}\}_{t\in[t_0,1]}$ and $\{x_{t_0,t}\}_{t\in[t_0,1]}$ starting in $\hat x_{t_0,t_0} = \xi$ and $x_{t_0,t_0} = \xi$, such that their velocity fields are $\hat v_t(\xi)$ and $v_t(\xi)$ respectively.
    Moreover, both $\hat x_{t_0,t}$ and $x_{t_0,t}$ are continuously differentiable in $\xi$, $t_0$ and $t$.
\end{assumption}

\begin{assumption}[Spatial Lipschitzness]\label{ass:spatial.lipschitz}
    The approximate velocity field $\hat v_t(x)$ is differentiable in both $x$ and $t$. Also, for each $t\in(0,1)$ there exists a constant $L_t$ such that $\hat v_t(x)$ is $L_t$-Lipschitz in $x$.
\end{assumption}
Assumption \ref{ass:flow.smoothness} is required since the flow matching method relies on solving the ODE \eqref{eq:triangular.velocity},
while Assumption \ref{ass:spatial.lipschitz} imposes smoothness constraints on the estimated velocity field.
From these two assumptions, it is already sufficient to provide a guarantee on the estimated joint distribution; see Theorem 1 of \cite{benton2023error}.

We now extend this result from the joint sampler to the posterior sampler.
The following theorem ensures proximity between the true posterior and its estimate from the sampler.

\begin{theorem}[Posterior Sampler Accuracy]\label{thm:posterior.W2.bound}
    Assume that the learned velocity field is constrained in the form of \eqref{eq:triangular.velocity}.
    Under Assumptions \ref{ass:flow.smoothness} and \ref{ass:spatial.lipschitz}, when $\mathcal L_{\text{FM}}(\hat v, v) \leq \varepsilon_N^2$, we have
    \[
    \mathbb E_{\pi_Y} \big [ W_2^2(\hat\pi(\theta\mid Y), \pi(\theta\mid Y)) \big ] \leq 2 Le^{2L} \varepsilon_N^2,
    \]
    where $L = \int_0^1 L_t \diff t$, and the expectation is taken over the marginal distribution of $Y$, denoted by $\pi_Y$.
\end{theorem}
Intuitively, the quality of the posterior sampler relies on two aspects: the extent that the objective function is minimized, and the spatial-Lipschitz constant of the true underlying velocity field.
The sampling error, measured by the 2-Wasserstein distance, grows exponentially with the Lipschitz constant $L_t$.

We note that $L_t$ also captures the variability of the $y$-component of velocity with respect to $\theta$, and the variability of the $\theta$-component of velocity with respect to $y$.
This inspires us to apply rescaling on the joint space $\mathcal Y \times \Theta$, such that each dimension has similar variance.
In fact, our experiments also support this insight.
With an extra scaling step applied on the joint space while training the neural network, the sampler yields better performance.
See Appendix \ref{sec:scaling} for further details.

\subsection{Consistency of Posterior Sampler}\label{sec:consistency}

An important factor in the upper bound in Theorem \ref{thm:posterior.W2.bound} is the control on the flow matching objective function, $\mathcal L_{\text{FM}}$.
We might be curious about the possibility to obtain $\mathcal L_{\text{FM}}\rightarrow 0$ as we increase the generative sample size $N$.
Theorem \ref{thm:MLP.consistency} provides an affirmative answer to this question, and establishes an asymptotic consistency result for the posterior sampler.
In addition to the existing assumptions, it relies on the following regularity conditions.

\begin{assumption}[Boundedness]\label{ass:farrell.1}
    There exists an absolute constant $M>0$ such that for all $t\in[0,1]$: (1) each entry of the true underlying velocity function $v_t$ satisfies $\sup_{t\in[0,1]} |(v_{t})_k|_\infty \leq M$ in all dimensions $1\leq k\leq d$, (2) the velocity learned by the feedforward neural network satisfies $\sup_{t\in[0,1]}|(\hat v_{t})_k|_\infty \leq 2M$ in all dimensions $1\leq k\leq d$, and (3) the joint samples are created from compact domains, $\mathcal Y \subset [-M,M]^n$, $\Theta \subset [-M,M]^d$.
    Here, $|\cdot|_\infty$ denotes the supremum norm of a univariate function.
\end{assumption}

\begin{assumption}[Smoothness]\label{ass:farrell.2}
    Assume that there exists $\beta \in \mathbb{N}_+$ such that the velocity field $v$ lies in the Sobolev ball $\mathcal W^{\beta, \infty}([0,1]\times \mathcal X)$.
\end{assumption}
Assumptions \ref{ass:farrell.1} and \ref{ass:farrell.2} are originally imposed by \cite{farrell2021deep} for the guarantee of functional estimation via the multilayer perceptron with ReLU activation.
These boundedness assumptions are fairly standard in nonparametrics (see \cite{farrell2021deep}), and the choice of $M$ in Assumption \ref{ass:farrell.1} may be arbitrarily large; no properties of $M$ are required apart from being finite.
The Sobolev smoothness in assumption \ref{ass:farrell.2} is an increment from the differentiability in $x$ and $t$ in Assumption \ref{ass:flow.smoothness}.
This increment is natural, and the additional requirement is that the velocity field does not vary too drastically with $x$ and $t$.
Detailed discussion of Sobolev spaces can be found in \cite{gine2021mathematical}.

\begin{theorem}[Consistency of Posterior]\label{thm:MLP.consistency}
    Under the regularity conditions given in Assumptions \ref{ass:farrell.1} and \ref{ass:farrell.2}, the velocity field $\hat v_t(x)$ learned by the multilayer perceptron with ReLU activation function achieves
    \[
    \mathcal L_{\text{FM}}(\hat v, v) \rightarrow 0 \quad \text{a.s.}
    \]
    when $N\rightarrow\infty$.
    As a result, if Assumptions \ref{ass:flow.smoothness} and \ref{ass:spatial.lipschitz} hold in addition, we have
    \[
    \sup_{y^*\in\mathcal Y}W_2(\hat\pi_{\theta\mid y^*}, \pi_{\theta\mid y^*}) \rightarrow 0
    \]
    almost surely for the generated data $\{y_i, \theta_i\}_{i=1}^N$.
\end{theorem}

\subsection{Inference Theory}
With the monotonicity constraint applied on the velocity field $v_t(x)$, any intermediate map $T_t$ is also monotone in $x$.
We can thus construct the following corollary of Theorem \ref{thm:MLP.consistency}, verifying that the credible sets $\hat C_\tau(y^*)$ defined in \eqref{eq:credible.set} converge to the oracle sets (also the MK depth regions) $C_\tau(y^*)$, defined in \eqref{eq:oracle.set}.
The proximity is measured by the Hausdorff distance,
\[
d_H(A,B) = \max\left\{ \sup_{a\in A}\inf_{b\in B} \|a-b\|, \sup_{b\in B}\inf_{a\in A}\|a-b\| \right\}.
\]

\begin{corollary}[Consistency of Credible Sets] \label{cor:credible.set.consistency}
    Assume that the true velocity field $v_t(x)$ is monotone in $x$, and all assumptions in Theorem \ref{thm:MLP.consistency} hold.
    Then almost surely for the generated data $\{y_i, \theta_i\}_{i=1}^N$, for any $\tau\in(0,1)$, as $N\rightarrow \infty$,
    \[
    \sup_{y^* \in \mathcal Y} d_H(\hat C_\tau(y^*), C_\tau(y^*)) \rightarrow 0.
    \]
\end{corollary}
Corollary \ref{cor:credible.set.consistency} shows that the posterior credible sets are consistent to the Monge-Kantorovich depth regions.
In terms of the inverse maps, consistency result on the MK conditional vector rank in \eqref{eq:MK.rank} can be constructed as an equivalent of Corollary \ref{cor:credible.set.consistency}.
Consider the estimated inverse map, $\hat R_{\theta\mid y^*} = [\hat G(y^*,\cdot)]^{-1}$.
For any $\theta\in\Theta$, denote the estimated MK conditional rank function as $\hat r_{\theta\mid y^*}(\theta) = \|\hat R_{\theta\mid y^*}(\theta) \|$.
The following result shows that this function acts like a p-value that compares $\theta$ with a posterior distribution.
\begin{proposition}[MK Conditional Rank]\label{prop:MK.conditional.rank}
    Under all assumptions in Corollary \ref{cor:credible.set.consistency}, with $\theta \sim \pi(\cdot \mid y^*)$, for any $\alpha \in (0,1)$ we have
    \[
    \pi_{\theta\mid y^*}(\hat r_{\theta\mid y^*}(\theta) > 1-\alpha) \rightarrow \alpha
    \]
    uniformly in $y^*\in\mathcal Y$, almost surely for the generated data $\{y_i, \theta_i\}_{i=1}^N$ as $N\rightarrow \infty$.
\end{proposition}

It is worth pointing out that the theoretical analysis in this section is more general than the specific Algorithm \ref{alg:triangular-flow-posterior} that we use in the experiment.
For any flow matching algorithm that is based on a block-triangular velocity field in \eqref{eq:triangular.velocity}, under the technical assumptions, the results in this section should hold.
More structural constraints can be imposed on the velocity field, e.g. various methods to ensure monotonicity, and we still have the consistency guarantee.
For the asymptotic results, the convergence rate with respect to $N$ might vary under different experiment setup and network structures.
It might be interesting to explore the finite-sample rates of these consistency results, and we leave those for further study.

\section{EXPERIMENTS}\label{sec:main-exp}
In this section, we empirically evaluate our approach from five complementary perspectives. Additional experiments and detailed setup are deferred to Appendix~\ref{app:additional_exp}.

\begin{figure}
    \centering
    \includegraphics[width=\linewidth]{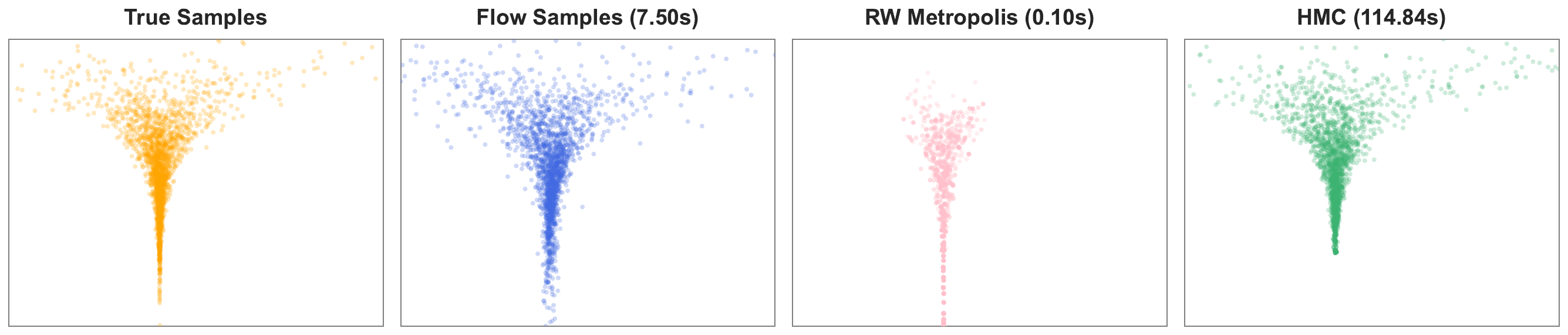}
    \caption{The traditional MCMC methods fail to search the entire region. The sampling time for each method is noted in the respective plot title. Our method requires one-time training (7.5 s); per sampling cost is nearly instantaneous and the model captures the underlying geometry.}
    \label{fig:neals-funnel}
\end{figure}

\paragraph{Flexibility of Flow Matching.}
We first examine the consistency of joint samplers. Sampling difficulties
arise when dealing with geometries exhibiting strong heteroskedasticity, such as \emph{Neal's Funnel} \citep{neal2003slice}:
\begin{align}\label{eq:neals.funnel}
\nu &\sim \mathcal{N}(0, 3^2),\quad x|\nu \sim \mathcal{N}(0, e^\nu).
\end{align}
Standard Monte Carlo Markov Chain (MCMC) methods like Hamiltonian Monte Carlo (HMC) often struggle with Neal's Funnel due to its extreme scale variation on $x$-axis; tight ``necks'' require small steps, while wide ``mouths'' favor large ones.
A fixed step size leads to poor mixing and slow convergence.
In contrast, generative models learn global transport maps that adapt across the geometry, enabling efficient sampling without region-specific tuning.
The timing annotations in Figure~\ref{fig:neals-funnel} highlight the low per-sample cost of our generative approach---flow model requires one-time training (7.5s), and the sampling is nearly free; however, for MCMC methods, we need to run the entire chain every time we sample. We also point out that both MCMC methods fail to cover the entire data space where Random Walk Metropolis Hastings (RWM) fails to cover the ``mouth'' and Hamiltonian Monte Carlo (HMC) fails to cover the ``neck''.

\paragraph{Scalability with Data Dimension $n$.}

\begin{figure}
    \centering
    \includegraphics[width=\linewidth]{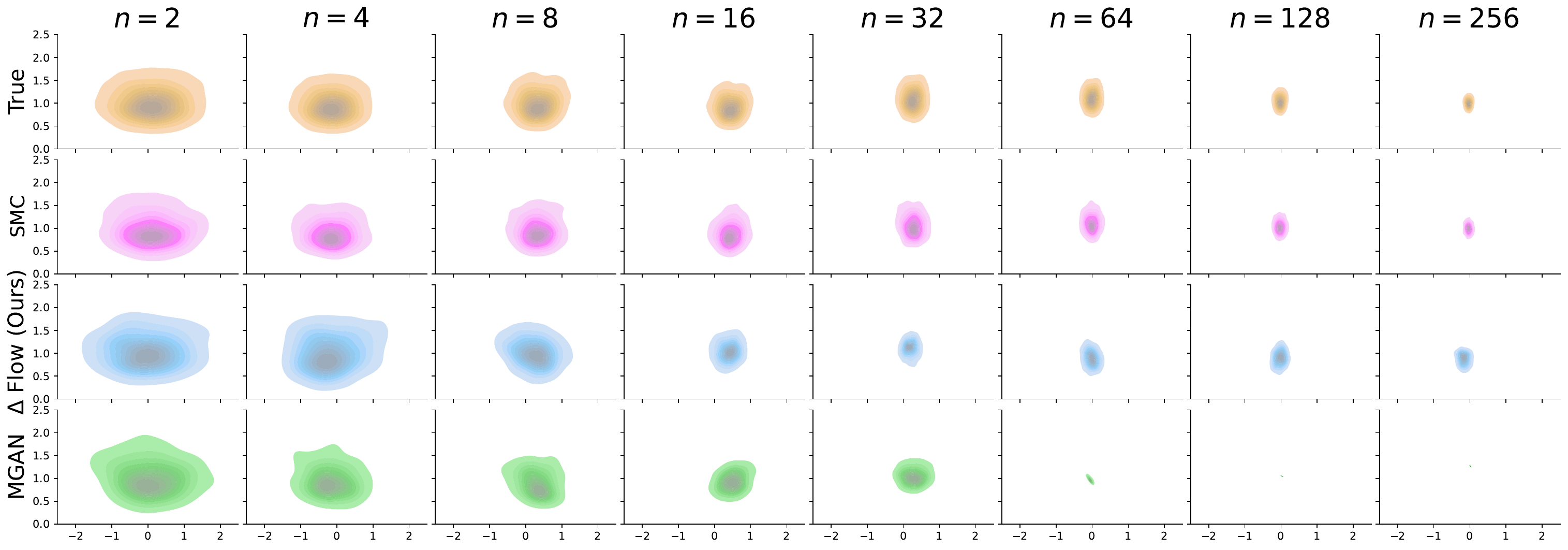}
    \caption{Posterior estimation for Gaussian conjugate model.
    Our model scales better than M-GAN with the increasing number of observations ($n \geq 64$).}
    \label{fig:guassian-conjugate-without-ss}
\end{figure}

In the second experiment, we show how our model scales with an increasing data dimension $n$ under simple Gaussian conjugate model (see Figure~\ref{fig:guassian-conjugate-without-ss}). Consider a one-dimensional Gaussian random variable $X$ with parameter $\mu$ and $\sigma^2$, i.e. $X\mid \mu, \sigma^2 \sim N(\mu, \sigma^2)$. For the conjugate relationship, we assume the prior distribution $\mu \mid \sigma^2 \sim N(\mu_0, \sigma^2/\kappa)$, and $\nu_0\sigma_0^2/\sigma^2 \sim \chi^2(\nu_0)$. For each $\mu_i, \sigma_i^2$, we generate $n$ many samples $x_{ij}$, $j \in [n]$, and we repeat the procedure $N$ times. We are given $X \in \mathbb{R}^{N \times n}$ as observed samples and $N$ pairs of parameters $\{ \mu_i, \sigma_i^2\}_{i=1}^N$. Our model tracks the posterior distribution without overshrinkage, achieving performance comparable to that of Sequential Monte Carlo (SMC), which directly uses the likelihood.
In contrast, the posterior estimated by M-GAN collapses starting from $n = 64$.

\paragraph{One-Time Training, Amortized Inference.}\label{app:computation}
We report computation times in Table~\ref{tab:time_comparison}, comparing our method ($\Delta$-Flow) with guided flow (G-Flow) \citep{zheng2023guided}, M-GAN \citep{baptista2024conditional}, rejection ABC (R-ABC), and Sequential Monte-Carlo ABC (S-ABC).

Flow matching methods (both $\Delta$-Flow and guided flow) train substantially faster than adversarial approaches such as M-GAN. 
Their training time is longer than ABC, but this cost is amortized: once trained, generative models ($\Delta$-Flow, G-Flow, and M-GAN) generate new posterior samples at negligible cost.
In contrast, ABC methods incur repeated simulation expense for every new observation, as the inference procedure must be rerun from scratch. See Appendix~\ref{sec:experiment-details} for dataset descriptions and further details.

\begin{table}
\centering
\resizebox{0.49\textwidth}{!}{
\begin{tabular}{@{}ccccccc@{}}
\toprule
\textbf{Dataset}                   & \textbf{Type}      & $\boldsymbol{\Delta}$\textbf{-Flow} & \textbf{G-Flow} & \textbf{M-GAN}  & \textbf{R-ABC}           & \textbf{S-ABC}           \\ \midrule
                                   & Training  & 719.4                                                & 512.55          & 4845.55         & \cellcolor[HTML]{C0C0C0} & \cellcolor[HTML]{C0C0C0} \\
\multirow{-2}{*}{Gaussian Linear}  & Inference & \textless{}0.01                                      & \textless{}0.01 & \textless{}0.01 & 0.25                     & 9.85                     \\
                                   & Training  & 818.4                                                & 504.88          & 4653.37         & \cellcolor[HTML]{C0C0C0} & \cellcolor[HTML]{C0C0C0} \\
\multirow{-2}{*}{Gaussian Mixture} & Inference & \textless{}0.01                                      & \textless{}0.01 & \textless{}0.01 & 0.37                     & 22.12                    \\
                                   & Training  & 849.7                                                & 997.81          & 4480.55         & \cellcolor[HTML]{C0C0C0} & \cellcolor[HTML]{C0C0C0} \\
\multirow{-2}{*}{Bernoulli GLM}    & Inference & \textless{}0.01                                      & \textless{}0.01 & \textless{}0.01 & 4.88                     & 20.07                    \\
                                   & Training  & 617.3                                                & 835.05          & 4792.71         & \cellcolor[HTML]{C0C0C0} & \cellcolor[HTML]{C0C0C0} \\
\multirow{-2}{*}{Two Moons}        & Inference & \textless{}0.01                                      & \textless{}0.01 & \textless{}0.01 & 0.40                     & 18.49                    \\
                                   & Training  & 906.1                                                & 509.87          & 6088.87         & \cellcolor[HTML]{C0C0C0} & \cellcolor[HTML]{C0C0C0} \\
\multirow{-2}{*}{SLCP}             & Inference & \textless{}0.01                                      & \textless{}0.01 & \textless{}0.01 & 6.5                      & 10.0                     \\ \bottomrule
\end{tabular}}
\caption{
Computation time (seconds) across SBI benchmarks.
For generative methods ($\Delta$-Flow, G-Flow, M-GAN), the table reports one-time training is costly but subsequent sampling is essentially free. 
For ABC methods, the reported time corresponds to generating posterior samples for each observation, which must be repeated for each inference task.
}
\label{tab:time_comparison}
\end{table}

\paragraph{Recovery of Posterior.}
Section~\ref{sec:theory} established in theory the consistency of posterior; we now demonstrate this empirically.
To quantitatively assess the ability of our sampler to recover the true posterior, we employ the classifier-based two-sample test (C2ST,  \cite{lueckmann2021benchmarking, lopez2016revisiting}), which trains a classifier to discriminate between samples from the true and estimated posterior distributions. 
A classification accuracy approaching 0.5 indicates ideal sampling performance.

Table~\ref{tab:posterior_benchmarks} compares our method ($\Delta$-Flow) against various baselines including rejection ABC (R-ABC), 
SMC-ABC (S-ABC) \citep{beaumont2009adaptive}, 
M-GAN \citep{baptista2024conditional},
guided flow (G-Flow) \citep{zheng2023guided},  \cite{kerrigan2024dynamic} and \cite{wildberger2023flow}.

Our proposed method achieves competitive performance across all five SBI benchmarks. 
It consistently outperforms the conditional OT baseline of \citet{kerrigan2024dynamic}.
This might be due to the additional flexibility brought by first estimating joint velocity field over $\mathcal Y\times \Theta$.
See Appendix~\ref{sec:comparison.to.concurrent} for more detailed discussion.

\begin{table*}
\centering
\resizebox{\linewidth}{!}{
\begin{tabular}{lccccccc}
\toprule
\textbf{Dataset} & $\boldsymbol{\Delta}$\textbf{-Flow} & \textbf{G-Flow} & \textbf{Kerrigan} & 
\textbf{Wildberger} & \textbf{M-GAN} & 
\textbf{R-ABC} & \textbf{S-ABC} \\
\midrule
Gaussian Linear   & \underline{0.71} & \textbf{0.69} & 0.89 & 0.97 & 0.85  & 
0.80 & 0.73 \\
Gaussian Mixture  & 0.82 & 0.85 & 0.96 &  0.57 &  \underline{0.73}  & 
0.80  & \textbf{0.65}\\
Bernoulli GLM     & 0.92 & 0.88 & 0.99 & 0.61 &  \underline{0.84}  & 
0.92  & \textbf{0.80} \\
Two Moons         & 0.74 & 0.78 & 0.99 & - &  \underline{0.67} & 
\textbf{0.64}  & 0.70 \\
SLCP              & \underline{0.93} & \textbf{0.91} & 0.97 & 0.96 & 0.98  & 
0.97 & 0.98\\
\bottomrule
\end{tabular}
}
\caption{C2ST metrics of posterior samplers across SBI benchmark datasets \citep{lueckmann2021benchmarking}.
Values closer to 0.5 indicate better performance.
The best value in each row is bolded and the second best is underlined.}
\label{tab:posterior_benchmarks}
\end{table*}

When compared to GAN-based posterior samplers (M-GAN), the results are mixed: $\Delta$-Flow outperforms M-GAN on Gaussian Linear and SLCP whereas M-GAN performs better on Gaussian Mixture, Bernoulli GLM and SLCP.
A key advantage of $\Delta$-Flow is that its training objective is simple ($\ell_2$ loss), avoiding adversarial optimization. As a result,  $\Delta$-Flow trains significantly faster than M-GAN (See Table~\ref{tab:time_comparison}).

\paragraph{Credible Set with Monotonicity.} We revisit the Gaussian conjugate example to illustrate the effect of enforcing monotonicity in the block-triangular flow.
As outlined in Section~\ref{sec:monotonicity}, monotonicity allows us to generate not only posterior samples but also well-defined Bayesian credible sets. To implement this, we use an input convex neural network (ICNN) to parameterize the monotone components of the map (see Appendix~\ref{app:monotonicity}).

Figure~\ref{fig:credible_main} summarizes the results: panel (a) illustrates the nested $\tau$-level credible sets produced by the flow, while panel (b) evaluates their calibration by comparing the empirical posterior mass with the nominal level $\tau$. 
These results demonstrate that the credible sets are not only visually well-behaved contours, but also achieve the expected nominal coverage under the posterior distribution.

\begin{figure}
    \centering
    \begin{subfigure}{0.2\textwidth}
    \includegraphics[width=\linewidth]{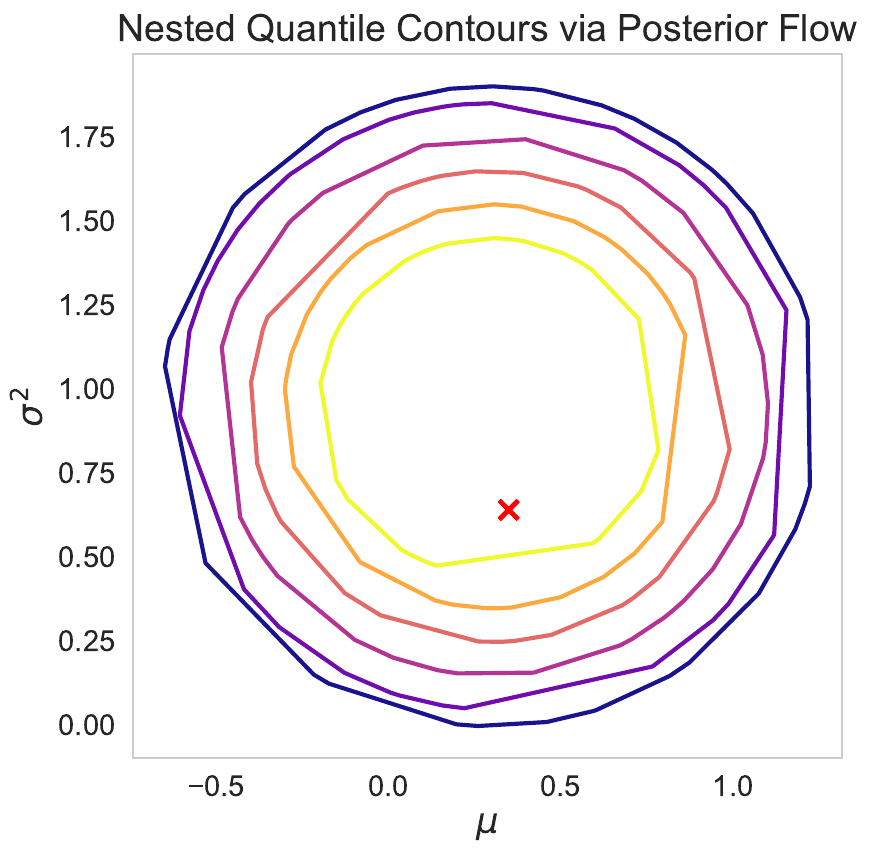}
    \caption{The $\tau$ credible sets}
    \label{fig:credible_sets}
    \end{subfigure}
    \begin{subfigure}{0.25\textwidth}
    \includegraphics[width = \linewidth]{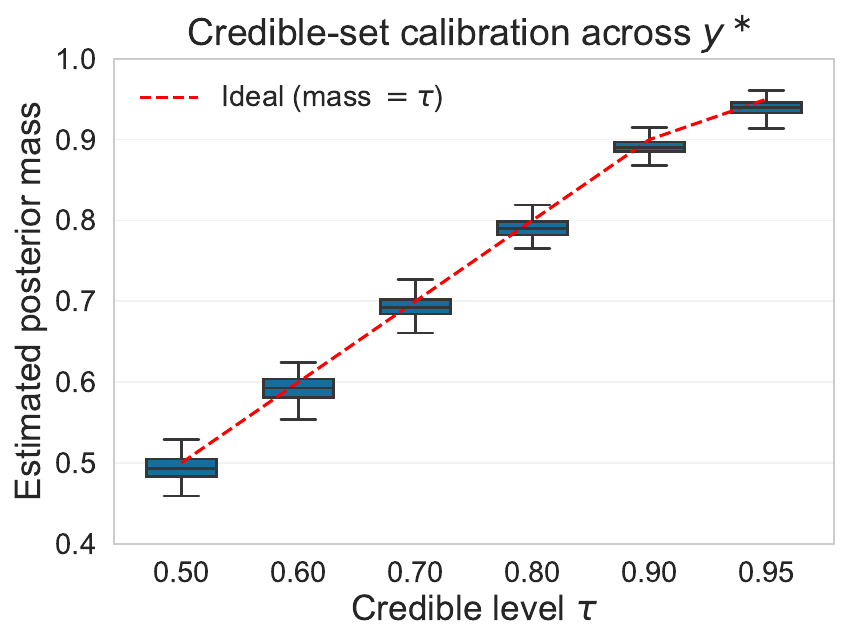}
    \caption{Coverage}
    \label{fig:credible_coverage}
    \end{subfigure}
    \caption{Panel (a) shows the credible sets with varying $\tau$ level ($\tau \in \{0.5, 0.6, 0.7, 0.8, 0.9, 0.95\}$) for gaussian conjugate model with $n=8$.
    Panel (b) shows the coverage rate of $\tau$-level credible set across 100 simulations.
    The red dashed line denotes the ideal $y=x$.
}
    \label{fig:credible_main}
\end{figure}

\begin{figure}
    \centering
\includegraphics[width=\linewidth]{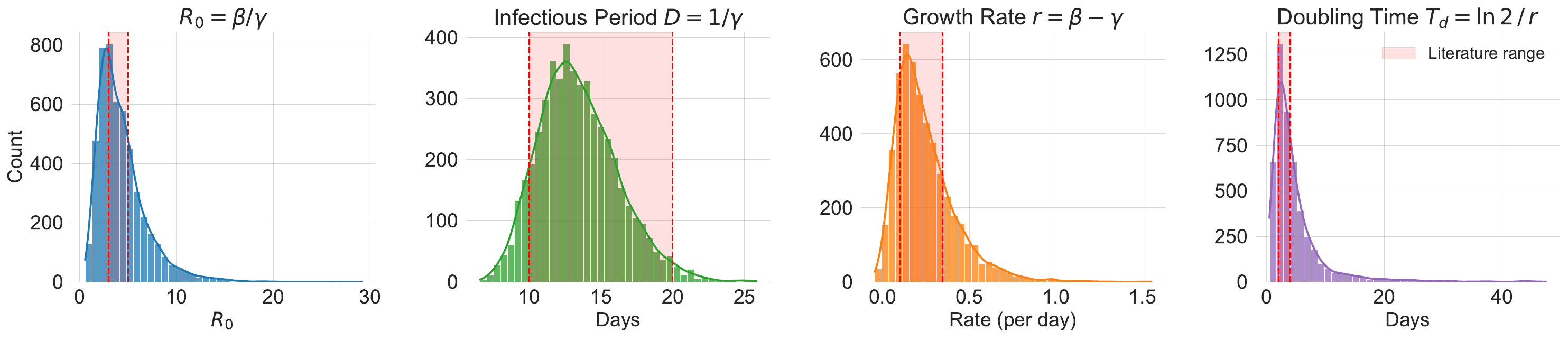}
\caption{Posterior distributions of four key epidemiological quantities derived from the SIR model fitted to 200-day Illinois COVID-19 data. }
    \label{fig:covid-19-posterior-stat}
\end{figure}

\paragraph{Real Data Application: COVID-19 Example.}
As a real-world data analysis, we used the COVID-19 surveillance data from the Illinois Department of Public Health, which contains daily case, death, and testing counts for 200 days.
From these observed quantities, we reconstruct susceptible (S), infected (I), and recovered (R) trajectories \citep{alqadi2022incorporating}, and fit a standard SIR model with $\beta$ and $\gamma$ being the transmission and removal rates, respectively.
We impose a log-normal prior on $(\beta,\gamma)$, sample its posterior using our proposed method, and construct the following quantities of interest:
(1) basic reproduction number $R_0 = \beta/\gamma$,
(2) days of infection $1/\gamma$,
(3) early exponential growth rate $r = \beta-\gamma$, and
(4) the doubling time $T_d = \log(2)/r$.
The experiment setup and detailed analysis are deferred to Appendix~\ref{covid-analysis}.
The Maximum A Posteriori (MAP) estimates yield an infection period of 14.6 days, and basic reproduction number ($R_0$) around 3.5, both consistent with published estimates.
Figure~\ref{fig:covid-19-posterior-stat} displays the posterior distributions of
these four epidemiologically meaningful functionals, with literature reference
ranges overlaid for comparison.

Further analysis on the effect of prior on Susceptible-Infected-Recovered (SIR) Model in epidemiology can be found in Appendix \ref{app:varying_prior}.

\section{CONCLUDING REMARKS}

This paper develops a simulation-based inference method to sample from a multivariate posterior distribution through flow matching.
Based on the posterior sampler, our method also provides inferential tools to implement uncertainty quantification, including the construction of Bayesian credible sets and testing the extent of a certain parameter value deviating from the posterior.
To apply these inferential tools, our method only requires training the map once.
This saves computation time compared to MCMC or ABC, which usually require re-runs with new observations.

One limitation of our method is the reduced performance when the observation dimension ($n$) is very high.
A possible, yet challenging, remedy is to replace the raw data with carefully chosen summary statistics and to train the flow matching model in this lower-dimensional latent space.
Another open direction is the extension to high-dimensional parameter spaces, where incorporating sparsity or structural correlation assumptions may be necessary for tractable learning.
Moreover, similar to other neural network-based approaches, our method is sensitive to hyperparameter tuning and requires careful implementation choices.
In practice, this means that for each dataset, the model should be properly tuned to achieve its best performance.
We leave these challenges for future work.

More broadly, our method shares a limitation that is common in push-forward generative models: difficulty with highly multimodal posteriors \citep{salmona2022can} and extremely heavy-tailed distributions \citep{wiese2019copula, tam2025statistical}.
Our Neal's Funnel experiment shows that moderate heavy-tailed behavior is recoverable, but robustness across all such regimes is not guaranteed.
We also leave these challenges for future work to solve.

\subsection*{Acknowledgment}
This research was supported in part through the computational resources and staff contributions provided for the Mercury high performance computing cluster at The University of Chicago Booth School of Business which is supported by the Office of the Dean.
Veronika Ro\v{c}kov\'{a}'s work is partially supported by NSF/DMS 2515709.

\clearpage
\bibliography{main}

\clearpage
\appendix

\onecolumn
\aistatstitle{Supplementary Materials}

\section{MATHEMATICAL NOTATION}\label{app:notation}

In this section we collect and define mathematical notation used throughout the paper. 

We denote the Euclidean norm of a vector $ x \in \mathbb{R}^d $ by $\|x\| = \sqrt{\sum_{i=1}^d x_i^2}$, and the Frobenius norm of a matrix $A \in \mathbb{R}^{d \times d}$ by $\|A\|_F = \sqrt{\sum_{i,j} A_{ij}^2}$. 
For a function $f: \mathbb{R}^d \to \mathbb{R}$, we denote the gradient by $\nabla f(x) = \left( \frac{\partial f}{\partial x_1}, \dots, \frac{\partial f}{\partial x_d} \right)^\top$, and the Hessian by $\nabla^2 f(x) \in \mathbb{R}^{d \times d}$, with entries $\left[ \nabla^2 f(x) \right]_{ij} = \frac{\partial^2 f}{\partial x_i \partial x_j}$.

For a vector-valued function $v: \mathbb{R}^d \to \mathbb{R}^d$, the Jacobian matrix is denoted by $ J_v(x) = \nabla v(x) \in \mathbb{R}^{d \times d} $, with entries $\left[ J_v(x) \right]_{ij} = \frac{\partial v_i}{\partial x_j}$. The divergence of $v$ is defined as the trace of the Jacobian
\begin{equation*}\label{eq:divergence}
 \nabla \cdot v(x) = \sum_{i=1}^d \frac{\partial v_i}{\partial x_i} = \text{tr}(\nabla v(x)).   
\end{equation*}
This quantity measures the net rate at which probability ``flows out" of a point under the vector field $v$, and appears in the continuity equation that governs time-dependent probability densities.

For a measurable map $T: \mathbb{R}^d \rightarrow \mathbb{R}^d$ and a probability measure $\mu$ on $ \mathbb{R}^d$, the pushforward measure $T_\sharp \mu$ is defined as the distribution of the random variable $T(X)$ when $X \sim \mu$. Formally, for any Borel set $A \subset \mathbb{R}^d$, we define
\begin{equation*}\label{eq:pushforward}
T_\sharp \mu (A) := \mu(T^{-1}(A)).    
\end{equation*}

The Wasserstein-2 distance between two probability measures $\mu$ and $\nu$ on $\mathbb{R}^d$ is defined as
\begin{equation*}\label{eq:wasserstein-2-distance}
W_2^2(\mu, \nu) = \inf_{\gamma \in \Gamma(\mu, \nu)} \int_{\mathbb{R}^d \times \mathbb{R}^d} \|x - y\|^2 \diff \gamma(x, y),    
\end{equation*}
where $\Gamma(\mu, \nu)$ is the set of all couplings (joint distributions) with marginals $\mu $ and $\nu$.

The Hausdorff distance between sets $A, B \subset \mathbb{R}^d$ is defined as
\begin{equation*}\label{eq:hausdorff}
 d_H(A, B) = \max \left\{ \sup_{a \in A} \inf_{b \in B} \|a - b\|, \sup_{b \in B} \inf_{a \in A} \|a - b\| \right\}.   
\end{equation*}

\section{RELATED WORKS}\label{app:related_works}

Bayesian posterior sampling has been addressed through a variety of generative methods. 
Here we review relevant approaches categorized by underlying technique.
 
\paragraph{Normalizing Flows.}
A normalizing flow (NF) is a sequence of invertible transformations that maps a base distribution $p_0(x)$ (e.g., a simple Gaussian) to a more complex target distribution $p_1(x)$.
The probability density function (PDF) is computed using the change of variables formula.
Let $f: \mathbb{R}^d \to \mathbb{R}^d$ be an invertible transformation with Jacobian determinant $|\det J_f(x)|$, then the density transformation follows
$$
p_1(x) = p_0(f^{-1}(x)) \left| \det J_{f^{-1}}(x) \right|,
$$
where $J_f(x) = \frac{\partial f}{\partial x}$ is the Jacobian matrix of the transformation.
One example is the real-valued non-volume preserving transformation (real NVP) \citep{dinh2017density}, which introduces block-triangular coupling layers to ensure tractable computation of the Jacobian determinant. In each layer, the input is split into two subsets, where one subset remains unchanged while the other is transformed conditionally. Specifically, affine transformations of the form
$$
x'_{1:d} = x_{1:d}, \quad x'_{d+1:D} = x_{d+1:D} \odot \exp(s(x_{1:d})) + t(x_{1:d}),
$$
yield a triangular Jacobian matrix whose log-determinant is a simple sum over outputs of $\mathbf{s}$, where $s(x_{1:d}) \in \mathbb{R}^{D-d}$ is a scale function, $t(x_{1:d}) \in \mathbb{R}^{D-d}$ is a translation function both parametrized by neural networks, and $\odot$ denotes an element-wise product.
The model provides exact likelihood computation via the change of variables formula; however, training is constrained by architectural choices that ensure invertibility (e.g., affine coupling layers).
 
Normalizing flows have been applied directly to simulation-based inference.
\citet{ardizzone2018analyzing} use invertible neural networks to learn the posterior in an amortized fashion, while \citet{radev2020bayesflow} propose BayesFlow, which couples a summary network with a conditional normalizing flow to perform amortized Bayesian inference from simulated data.
Both methods condition on static observations and require invertible architectures with tractable Jacobians, whereas our approach avoids invertibility constraints entirely by learning a velocity field via flow matching.

\paragraph{Continuous Normalizing Flow}
A Continuous Normalizing Flow (CNF) \citep{chen2018neural} generalizes NF to continuous-time dynamics by modeling the transformation as a solution to an ODE, $\frac{dx}{dt} = v_t(x_t, \omega)$, ($t \in [0,1]$) where $v_t(x_t,\omega)$ is the velocity field with the learnable parameters $\omega \in \Omega$, for example, by neural networks. The model still relies on the modified version of the change of variable formulas.
\begin{lemma}[Instantaneous Change of Variables \citep{chen2018neural}]
Let $x(t) \sim p_t$ be a finite continuous random variable, and $d x_t = v_t(x_t) dt$ be the velocity field. If $v$ is uniformly Lipschitz continuous in $x(t)$ and continuous in $t$, then the change in log probability follows a differential equation 
\begin{equation*}\label{eq:cnf}
    \frac{\partial p_t \big( x(t) \big) }{\partial t} = - \mathrm{tr} \Big ( \frac{ \diff v}{\diff x(t)} \Big )
\end{equation*}

\end{lemma}

\paragraph{Diffusion Model} Diffusion models \citep{song2020score} add noise to data (Forward Process) and learn to reverse this process via score estimation (Reverse Process). While powerful, they require computationally expensive training and sampling, and rely on accurate score estimates. The likelihood computation is possible through integration of the score function. The SDE-based diffusion model has a connection to the probability flow ODE relying on the Fokker--Planck equation. 

\begin{equation*}\label{eq:dp-forward-process}
\diff x_t = f_t(x_t) \diff t + g_t \diff B_t,
\end{equation*}
where $f_t(x_t)$ is a drift function, $g_t$ is the noise scaling, and  $B_t$ is a Brownian motion.

\begin{equation*}\label{eq:dp-reverse-process}
\diff x_t = [f_t(x_t) - g_t^2 \nabla_x \log p_t(x_t)] \diff t + g_t \diff \tilde{B}_t
\end{equation*}
where $\nabla_x \log p_t(x_t)$ is the score function. 

\begin{align*}\label{eq:sde-ode-equivalence}
\diff x &= f(t,x) \diff t + g(t) \diff B_t\\
\Leftrightarrow \frac{\partial p_t(x) }{\partial t} & = - \nabla_x (f p_t) + \frac{1}{2} g^2 \nabla_x ^2 p_t \quad \text{(Fokker--Planck)}\\
& = - \nabla_x (f p_t) + \frac{1}{2} g^2 \nabla_x (p_t \nabla_x \log p_t)\\ 
& = - \nabla_x \Big( (f- \frac{1}{2}g^2 \nabla_x \log p_t) p_t \Big)\\
\Rightarrow \diff x &= \tilde{f}(x,t) \diff t, \qquad \tilde{f}(x,t) = f(x,t) - \frac{1}{2}g(t)^2 \nabla_x \log p_t(x)
\end{align*}

\subsection{Guidance vs. Conditional Sampling} 

Guidance techniques are employed to steer the sampling process of diffusion models toward specific desired outcomes. Two prominent methods are classifier guidance \citep{song2020score} and classifier-free guidance \citep{ho2022classifier}. It starts with the simple score decomposition of the form
\begin{equation*}\label{eq:score-decomposition}
    \nabla \log p_t(x|y) = \nabla \log p_t(y|x) + \nabla \log p_t(x).
\end{equation*}

Introduced by \citet{song2020score}, the classifier guidance utilizes an external classifier, $p(y|x)$, trained to predict the class label from a given image. Note that the model is still trained marginally $p(x)$ and the classifier signal only plays a role in the sampling procedure. During sampling, the gradients from this classifier are combined with the diffusion model's score estimates to guide the generation process toward images that are more likely to belong to the desired class. This approach effectively biases the model to produce outputs aligned with specific class labels.

The classifier-free guidance (CFG) by \cite{ho2022classifier} eliminates the need for an external classifier by training a single diffusion model capable of both conditional and unconditional generation. During training, the model learns to handle inputs with and without conditioning information (e.g., class labels) by introducing the Bernoulli variable indicating the conditionality. At inference, the model interpolates between these two modes by adjusting the guidance scale. 
\begin{equation}\label{eq:CFG}
    \tilde{v}_t(x|y) = (1 - w) \cdot v_t(x|\emptyset) + w \cdot v_t(x|y)
\end{equation}
This interpolation (\ref{eq:CFG}) can be mathematically represented as a weighted combination of the conditional and unconditional score estimates.

Guided flow \citep{zheng2023guided} is the adaptation of classifier-free guidance (CFG) on flow matching from the diffusion setting. As in \cite{ho2022classifier}, the score decomposition strategy is used for time dependent classifiers:
\begin{equation*}
    \tilde{v}_t(x|y) = v_t(x) + b_t \cdot \nabla \log p_\phi(y | x, t)
\end{equation*}

The primary advantage of guidance methods lies in their ability to control the fidelity of generated samples with respect to the conditioning information.
However, we note that the goal of guidance is not exact conditional estimation; the guided distribution is a tempered approximation whose fidelity depends on the guidance scale.

\subsection{Posterior Sampling via Generative Models}
We now review generative approaches that directly target posterior distributions.

\paragraph{Generative Adversarial Network (GAN) based posterior sampling}
\cite{wang2023adversarialbayesiansimulation} and \cite{baptista2024conditional} leverage matching of joint samples of data and parameters against pairs of marginal data and noise.
\cite{baptista2024conditional} imposes additional constraints such as monotonicity and block-triangular structure, enhancing the stability of posterior estimation.
Our method shares the block-triangular framework with \cite{baptista2024conditional} but replaces adversarial learning with a flow matching objective, resulting in simpler training and faster convergence (see Table~\ref{tab:time_comparison}).
\citet{bendel2023regularized} build upon the conditional GAN framework \citep{mirza2014conditional} and propose a regularization scheme to enhance training stability and improve sample diversity, with a primary focus on linear inverse problems.

\paragraph{Diffusion based posterior sampling}
Diffusion-based posterior sampling, by \cite{chung2022diffusion}, incorporates classifier signals directly into training to handle general noisy inverse problems, avoiding explicit measurement consistency constraints. 
\begin{equation*}
    \frac{dx}{dt} = -\frac{\beta(t)}{2} x - \beta(t) \left( \nabla_x \log p_t(x) + \nabla_x \log p_t(y|x) \right) + \sqrt{\beta(t)} \diff \bar{B}_t
\end{equation*}
The posterior mean estimation via Tweedie formula for the likelihood approximation during the diffusion posterior sampling:
\begin{equation*}
    \hat{x}_0 := \mathbb{E}[x_0 | x_t] \approx \frac{1}{\sqrt{\bar{\alpha}(t)}} \left( x_t + (1 - \bar{\alpha}(t)) s^*_\theta(x_t, t) \right)
\end{equation*}

\paragraph{Flow based posterior sampling}
Several recent works apply flow models to posterior estimation.
\citet{benhamu2024dflowdifferentiatingflowscontrolled} and \citet{pokle2024trainingfreelinearimageinverses} propose training-free methods for conditional generation in linear inverse problems, modifying the generative process at inference time to enforce data consistency rather than learning a posterior distribution directly.
FlowDPS \citep{kim2025flowdps} adapts the diffusion posterior sampling framework of \citet{chung2022diffusion} to flow matching models, but continues to rely on score estimation via learned gradients of log densities.
In contrast, our method departs from score-based training entirely, using a structured velocity decomposition to enable exact posterior recovery.
 
\citet{wildberger2023flow} present a simulation-based framework using flow models for posterior estimation.
They employ forward simulations to generate training data and then train a conditional flow to approximate the posterior, conditioning on static observations~$y$.
This setup is conceptually close to ours, but methodologically distinct: \cite{wildberger2023flow} condition on the raw observation as a fixed input to the velocity network, whereas we learn a joint velocity field over $(\mathcal{Y} \times \Theta)$ in which both data and parameter components evolve over time.
By decomposing the joint transport into marginal and conditional components, our approach distributes the geometric complexity across two cooperative sub-problems rather than placing the entire representational burden on a single conditional network.

\subsection{Conditional Optimal Transport and Related Frameworks}\label{app:cot_related}
 
A number of concurrent and recent works approach conditional sampling through the lens of optimal transport, and we discuss their relationship to our method here.
 
\citet{kerrigan2024dynamic} generalize the Benamou--Brenier theorem to the conditional setting, learning a conditional velocity field that transports a source distribution to the target while keeping the conditioning variable~$y$ fixed (identity coupling in~$\mathcal{Y}$).
Their theoretical focus is on minimizing the conditional Wasserstein distance between source and target.
\citet{chemseddine2024conditional} introduce a conditional Wasserstein distance $W_{p,Y}$ and derive an OT Bayesian Flow Matching algorithm that penalizes mass transport in the conditioning variable's space, arriving at a comparable velocity target in the Euclidean setting.
Both methods learn a \emph{conditional} velocity of the form $\frac{d\theta_t}{dt} = g_t(\theta_t; y)$, where $y$ enters as a static input.
We note that under linear interpolation in Euclidean space with matched data marginals, the target velocity fields of these approaches and ours coincide; the methods differ in what is held fixed during training and in the theoretical guarantees that follow.
 
\citet{generale2024conditional} address a different goal: forecasting dynamic systems where data arrives as a time series.
They perform sequential Bayesian inference in which the source distribution for the flow at each step is the previous posterior approximation, not a fixed noise distribution.
\citet{isobe2024extended} propose an extension to conditional flow matching that addresses the problem of flowing between conditions (e.g., from $\pi(\theta \mid y_1)$ to $\pi(\theta \mid y_2)$) by integrating over a path in the conditional space, optimizing for smoothness of the mapping with respect to the conditioned variables.
Both works address fundamentally different inference tasks from ours.
 
\subsection{Transport-Based Methods for Bayesian Computation}\label{app:transport_related}
 
Several works leverage transport maps to accelerate or replace traditional Bayesian computation, without necessarily using flow matching.
 
\citet{makkuva2020optimal} learn static OT maps between two fixed distributions using ICNN parameterizations of convex potentials, with the goal of recovering a global Monge map via direct optimization.
In our approach, convexity is not required for posterior sampling itself but serves as an optional constraint for constructing credible sets (Section~\ref{sec:monotonicity}).
\citet{li2025optimal} cast posterior learning directly as an OT problem and estimate a transport map pushing the prior onto the posterior; in simple settings the OT formulation yields linear maps that act as efficient posterior samplers.
Our method shares the high-level goal of learning a deterministic prior-to-posterior map but obtains it implicitly through a flow matching objective rather than solving the OT problem directly.
 
\citet{duan2023transport} build randomized approximate transport maps to precondition MCMC (Transport Monte Carlo), improving asymptotic efficiency while still relying fundamentally on Monte Carlo chains.
Similarly, \citet{hoffman2019neutra} train a neural transport map to ``neutralize'' the geometry of the posterior and then run HMC on the transformed space.
In both cases, the transport map serves as an auxiliary preconditioner rather than a stand-alone sampler.
Our method produces posterior samples directly by integrating the learned velocity field, without requiring Hamiltonian dynamics or accept--reject mechanics.
 
\citet{katzfuss2024scalable} estimate a triangular Rosenblatt/Knothe transport map for high-dimensional spatial fields using Gaussian process regressions, maximin spatial ordering, and Bayesian regularization.
Although both their method and ours involve triangular transport ideas, the approaches differ substantially: we do not assume spatial structure or impose GP priors, and we learn the transport implicitly via flow matching rather than through structured GP regressions.
 
\citet{jiang2025simulation} address simulation-based inference via score matching combined with Langevin dynamics, targeting settings with high-dimensional parameter spaces.
\citet{polsonGenerativeAIBayesian2024} propose quantile-based deep generative samplers trained with pinball losses.
Our method differs from both in using flow matching to learn time-dependent conditional flows, and in its ability to handle multivariate posteriors with joint velocity decomposition.


\section{TECHNICAL REMARKS}\label{sec:technical.remarks}

\subsection{Comparison with Other Flow-based Methods}\label{sec:comparison.to.concurrent}

There are two notable concurrent works that also use flow matching to learn the posterior distribution.
\cite{chemseddine2024conditional} minimizes the conditional Wasserstein distance to learn the velocity field leading to posterior distribution, while \cite{kerrigan2024dynamic} generalized Benamou--Brenier optimization, leading to an equivalent optimal transport map toward the posterior.

\paragraph{Optimal Transport vs. Consistency.}
One key difference between these methods and ours is the main objective of learning the velocity field.
\cite{kerrigan2024dynamic} and \cite{chemseddine2024conditional} aim to minimize the conditional Wasserstein distance between the source distribution and the target posterior distribution.
Our method, on the contrary, is mainly driven by the theoretical guarantee like Theorem \ref{thm:MLP.consistency}, i.e., the consistency of the learned posterior distribution compared to the true underlying posterior, in terms of Wasserstein distance.
The theoretical motivation is different.

The method of \cite{kerrigan2024dynamic}, for instance, is theoretically constrained on the specific geodesic path in the conditional Wasserstein space.
Our method, on the other hand, allows for flexible interpolation schemes (linear, geodesic, or as in \cite{albergo2022building}), without affecting consistency guarantees of Section \ref{sec:theory}.
The consistency results only rely on the accuracy of the learned joint velocity field, rather than the specific interpolation path.

Our method also allows for the possibility to enforce monotonicity of the transport map using ICNN, leading to the Monge-Kantorovich conditional vector quantile map \citep{chernozhukov2017monge}.
This allows us to generate nested Bayesian credible sets and compute conditional vector ranks, inferential tools that are difficult to obtain with standard generative models.
On the contrary, enforcing such monotonicity in Kerrigan's framework would conflict with the objective of learning the true optimal transport path.
Note that this should not be seen as a comparative advantage of our method per se, but rather the result of a different emphasis in method design.
Our method trades off strict OT optimality to gain rigorous uncertainty quantification capabilities.

\paragraph{Learning the Posterior Directly vs. Learning the Joint First.}
Technically, the use of block-triangular structure is different for our method.
Methods like \cite{chemseddine2024conditional}, \cite{wildberger2023flow}, and \cite{kerrigan2024dynamic} directly consider the posterior velocity field over $\Theta$ space alone, with observation $y^*$ fixed.
Our proposed method, on the other hand, learns a joint, block-triangle velocity field over $\mathcal Y\times\Theta$.
The block-triangular map theory ensures that an accurately learned joint distribution will lead to an accurately learned posterior distribution.

Therefore, our approach allows for an extra layer of flexibility.
An important aspect of this flexibility is the \emph{decoupling of marginal transport from conditional transport}.
Fixed-$y^*$ methods use the form of velocity $\dot\theta_t = v_t(y^*, \theta_t)$, and learns the velocity as a function of the static, raw observation $y^*$.
This places a heavy representational burden on the network for learning $v_t$.
Our approach, on the contrary, considers both $\dot y_t = f_t(y_t)$, i.e. marginal transport, and $\dot \theta_t = g_t(y_t, \theta_t)$, which now depends on dynamic $y_t$.
That is to say, the task of learning posterior velocity field is now decomposed into two sub-problems.
By offloading the modeling of the data geometry to $f_t$, the conditional network $g_t$ only needs to learn a simpler mapping: how parameters relate to the flow of data, rather than how they relate to static $y^*$.
Theoretically, this allows the smoothness assumptions on $f_t$ and $g_t$ to be more mild.
By decomposing the joint transport into marginal and conditional components, we effectively distribute the geometric complexity of the target distribution across two simpler, cooperative functions, thus relieving representational burden.

Moreover, our proposed method allows the inference to be guided by the marginal transport $f_t$, rather than relying on the network to implicitly interpolate the geometry of $y$ from static training data points.
This may help in practice, especially when the conditioned observation $y^*$ falls into a low-likelihood area.
In our framework, the marginal velocity field $f_t$ acts as a structural guide, effectively bridging the geometric gaps between training data points.
This additional flexibility may explain improved empirical performance of our method over, e.g., \citet{kerrigan2024dynamic}, on several SBI benchmarks; see Table \ref{tab:posterior_benchmarks}.

\subsection{ICNN Representability Issue}\label{sec:ICNN.representation.issue}

In this subsection, we seek to elaborate the ICNN representability issue discussed in Remark \ref{rmk:ICNN.representation}.
Under the scheme \eqref{eq:ICNN.representation}, our method learns the velocity field
\[
    \hat g_t(y_t, \theta_t) = \nabla_{\theta_t} \hat \psi_t(y_t, \theta_t)
\]
by learning a convex function $\psi_t$ with ICNN.
It is indeed reducing representational capacity, in that not all monotone maps can be represented by a monotone velocity.
Specifically, under linear interpolation path, the velocity can be written as
\[
g_t(y,\theta \mid x_0,x_1) = G(y^*, \theta_0) - \theta_0.
\]
Indeed, learning a velocity $\hat g_t(y,\theta)$ monotone in $\theta$ ensures that $\nabla_\theta g_t \succeq 0$, which implies $\nabla G \succeq I$.
Therefore, those non-expanding monotone velocity fields cannot be represented.
However, we can practically mitigate this limitation by initializing the source distribution with a much smaller variance than the target.
This ensures the optimal transport map is naturally expansive ($\nabla G \succeq I$), bringing the problem within the representational capacity of the monotone velocity field.

In order to address the representational issue more fundamentally, we propose an alternative velocity formulation that allows for contractive maps.
We can instead learn the velocity field
\[
\hat g_t(y_t, \theta_t) = \nabla_{\theta_t} \hat \psi^\ast_t(y^*, \theta_t) - \theta_0,
\]
where $\hat \psi^\ast_t$ is a convex function learned by ICNN.
One technical issue is that flow matching requires $\theta_t$ as input of $\hat g_t$, instead of $\theta_0$.
To address this issue, we can replace it by $\theta_0 \approx \theta_t - t\hat g_t$ (under linear interpolation path).
The velocity field is therefore formulated as
\[
\hat g^\ast_t(y_t, \theta_t) = \frac{\nabla_{\theta_t} \hat \psi_t(y^*, \theta_t) - \theta_t}{1-t}.
\]
Under such formulation, it is possible for contracting monotone maps $G$ to be represented by such a velocity $g_t^\ast$.
Therefore this formulation avoids the loss of representation, and we can still learn the velocity by ICNN.

\section{PROOF OF THEORY IN SECTION \ref{sec:theory}}\label{app:proof}

\subsection{Proof of Theorem \ref{thm:posterior.W2.bound}}

Note that the true joint distribution $p_1$ can be recovered by passing $p_0$ through the map $T$ as in \eqref{eq:triangular.map}, and the joint velocity field learned from the neural network leads to an estimated map $\hat T$ that transports $p_0$ to $\hat p_1$, an estimated joint distribution.
Formally,
\[
\hat p_1 = \hat T_\sharp p_0, \quad p_1 = T_\sharp p_0.
\]
It is known that $\hat p_1$ is consistent to $p_1$ when the velocity is learned accurately enough in terms of the flow matching objective function $\mathcal L_{\text{FM}}$ given in \eqref{eq:flow-matching-loss-intractable}.
In fact, by Theorem 1 in \cite{benton2023error}, under Assumptions \ref{ass:flow.smoothness} and \ref{ass:spatial.lipschitz}, when $\mathcal L_{\text{FM}} \leq \varepsilon_N^2$, we have
\begin{equation}\label{eq:benton.theorem}
W_2(\hat p_1, p_1) \leq \varepsilon_N \exp\left\{\int_0^1 L_t \diff t\right\}.
\end{equation}

Recall that the estimated posterior $\hat \pi(\theta \mid y^*)$ is recovered by the partial map $\hat G(y^*,\cdot)$, which is a slice of the joint map $\hat T$.
Also note that the true underlying map $G(y^*, \cdot)$ recovers the true posterior $\pi(\theta\mid y^*)$, a consequence of Lemma \ref{lm:map.velo} and Theorem 2.4 in \cite{baptista2024conditional}.
Formally,
\[
\hat\pi_{\theta\mid Y=y^*} = \hat G(y^*, \cdot) _\sharp p_0^\Theta, \quad \pi_{\theta\mid Y=y^*} = G(y^*, \cdot) _\sharp p_0^\Theta.
\]
We shall now find the relation between $W_2(\hat p_1, p_1)$ and $W_2(\hat\pi_{\theta\mid Y}, \pi_{\theta\mid Y})$.
Under the block-triangular map framework \eqref{eq:triangular.map},
\begin{align*}
    W_2^2(\hat p_1, p_1) &= W_2^2(\hat T_\sharp p_0, T_\sharp p_0)\\
    &= \int \|\hat T(y_0,\theta_0) - T(y_0,\theta_0) \|^2 p_0(\diff y_0, \diff \theta_0)\\
    &= \int \| \hat F(y_0) - F(y_0) \|^2 p_0^\mathcal Y (\diff y) + \int \|\hat G(\hat F(y_0),\theta_0) - G(F(y_0),\theta_0)\|^2 p_0(\diff y_0, \diff \theta_0).
\end{align*}
Since both terms are nonnegative, the upper bound \eqref{eq:benton.theorem} implies that either term is upper bounded by the same rate.
By Assumption \ref{ass:spatial.lipschitz}, since $\hat v_t(x)$ is spatially Lipschitz, i.e. $\| \hat v_t(x) - \hat v_t(x')\| \leq L_t \|x-x'\|$, we obtain a similar Lipschitzness in the joint estimated map $\hat T$ by integrating over time.
As a slice of $\hat T$, this property is also shared by $\hat G$,
\[
\| \hat G(\hat F(y_0), \theta_0) - \hat G(F(y_0), \theta_0) \| \leq L\|\hat F(y_0) - F(y_0)\|.
\]
By triangular inequality, we get
\begin{equation*}
\| \hat G( F(y_0), \theta_0) -  G(F(y_0), \theta_0) \| \leq
\| \hat G( F(y_0), \theta_0) -  \hat G(\hat F(y_0), \theta_0) \| + \| \hat G( \hat F(y_0), \theta_0) -  G(F(y_0), \theta_0) \|.
\end{equation*}
Using $(a+b)^2 \leq 2a^2+2b^2$ and taking an expectation of the square over $p_0(y_0,\theta_0)$, we get
\begin{align*}
    &\int \| \hat G( F(y_0), \theta_0) - G(F(y_0), \theta_0) \|^2 p_0(\diff y_0, \diff \theta_0)\\
    &\leq 2W_2^2(\hat p_1, p_1) + (2L-2)\int\| \hat F(y_0) - F(y_0)\|^2 p_0^\mathcal Y (\diff y_0)\\
    &\leq 2Le^{2L}\varepsilon_N^2.
\end{align*}
Therefore, we arrive at
\begin{align*}
\int W_2^2(\hat\pi_{\theta\mid y}, \pi_{\theta\mid y}) \pi_Y(\diff y) &= \int \|\hat G(y,\theta_0) - G(y,\theta_0) \|^2 p_0^\Theta(\diff \theta_0) \pi_Y(\diff y)\\
&= \int \|\hat G(F(y_0),\theta_0) - G(F(y_0),\theta_0) \|^2 p_0(\diff \theta_0, \diff y_0)\\
&\leq 2L e^{2L} \varepsilon_N^2.
\end{align*}
This finishes the proof of Theorem \ref{thm:posterior.W2.bound}.

\subsection{Proof of Theorem \ref{thm:MLP.consistency}}

Although the way our method trains $\hat v_t(x)$ is through minimizing $\mathcal L_{\text{CFM}}$ in \eqref{eq:cfm-loss} instead of $\mathcal L_{\text{FM}}$ in \eqref{eq:flow-matching-loss-intractable}, since they both have equal gradients with respect to the neural network parameters $\varphi$ (see \eqref{eq:equivalence-fm-cfm}).
Indeed, Proposition 1 in \cite{albergo2022building} also verifies that
\[
v = \argmin_{\hat v} \mathcal L_{\text{FM}}(\hat v, v) = \argmin_{\hat v} \mathcal L_{\text{CFM}}(\hat v, v).
\]
It is interesting to point out that it is usually impossible for $\mathcal L_{\text{CFM}}$ to converge to zero.
Instead, it converges to a constant unrelated to $\hat v$,
\[
\int_t \E_{p_t(x\mid z), q(z)}\| v_t(x\mid z)\|^2\diff t - \int_t \E_{p_t(x)} \|v_t(x)\|^2 \diff t.
\]
There is generally no guarantee on this quantity.
Despite $\mathcal L_{\text{FM}}$ being intractable, we can still apply the theoretical framework of \cite{farrell2021deep}.
Note that the true underlying velocity $v$ is the minimizer of the alternative objective function $\mathcal L_{\text{CFM}}$, and the estimated velocity field $\hat v$ is obtained by empirically minimizing the empirical version of $\mathcal L_{\text{CFM}}$,
\[
\hat v = \argmin_{\hat v \in \mathcal V_{\text{MLP}}} \frac{1}{N}\sum_{i=1}^N \| \hat v_{t_i}(x_i) - v_{t_i}(x_i \mid z_i) \|^2,
\]
where $\mathcal V_{\text{MLP}}$ is the function class learnable by the multilayer perceptron, $t_i$ is the $i$-th sampled time point, and $x_i = (y_i,\theta_i)$ is the $i$-th generated sample.

By Theorem 1 in \cite{farrell2021deep}, there exists a constant $C>0$ unrelated to $N$ such that for each dimension $1\leq k\leq n+d$, with probability no less than $1-\exp\{-N^{\frac{\beta}{\beta+n+d}}\log^8 N\}$,
\[
\E_{t,x}(\hat v_{t,k}(x) - v_{t,k}(x))^2 \leq C\left(N^{-\frac{\beta}{\beta+n+d}}\log^8 N + \frac{\log\log N}{N}\right).
\]
Note that the expectation is taken over all variables of the functions $\hat v$ and $v$, i.e. $t\sim \text{Unif}(0,1)$ and $x \sim p_1$.
Adding all dimensions up, it is exactly the flow matching objective function $\mathcal L_{\text{FM}}(\hat v, v)$. That is,
\[
\mathcal L_{FM} \leq C(n+d)\left(N^{-\frac{\beta}{\beta+n+d}}\log^8 N + \frac{\log\log N}{N}\right) := \varepsilon_N^2,
\]
with probability no less than $1-(n+d)\exp\{-N^{\frac{\beta}{\beta+n+d}}\log^8 N\}$.
This upper bound provides a finite-sample rate for $\varepsilon_N^2$.
Moreover, observe that
\[
\sum_{N=1}^\infty \P(\mathcal L_{\text{FM}} > \varepsilon_N^2) \leq (n+d)\sum_{N=1}^\infty \exp\{-N^{\frac{\beta}{\beta+n+d}}\log^8 N\} <\infty.
\]
Then by the Borel-Cantelli's Lemma, we have $\mathcal L_{\text{FM}} < \varepsilon_N^2$ almost surely for the generated dataset.
By Theorem \ref{thm:posterior.W2.bound}, we have
\[
\E_{\pi_Y} W_2^2(\hat\pi_{\theta\mid Y}, \pi_{\theta\mid Y}) = O\left( N^{-\frac{\beta}{\beta+n+d}}\log^8 N\right),
\]
a rate that is polynomial to $N$.
This immediately translates to
\[
W_2(\hat\pi_{\theta\mid Y}, \pi_{\theta\mid Y}) \rightarrow_{\pi_Y} 0.
\]

To prove uniform convergence, first note that $\mathcal Y$ is compact by Assumption \ref{ass:farrell.1}.
Second, from Assumptions \ref{ass:flow.smoothness} and \ref{ass:spatial.lipschitz} we can also show that $W_2(\hat \pi_{\theta\mid y}, \pi_{\theta\mid y})$ is equicontinuous in $y$.
To see this, note that by triangular inequality, for any $y,y'\in\mathcal Y$,
\[
|W_2(\hat \pi_{\theta\mid y}, \pi_{\theta\mid y}) - W_2(\hat \pi_{\theta\mid y'}, \pi_{\theta\mid y'})| \leq W_2(\hat \pi_{\theta\mid y}, \hat \pi_{\theta\mid y'}) + W_2( \pi_{\theta\mid y}, \pi_{\theta\mid y'}).
\]
By the Lipschitzness assumption in the velocity field, the overall map is also Lipschitz.
The right hand side can be upper bounded by $C\|y-y'\|$ for some constants $C>0$.
Therefore, the 2-Wasserstein distance $W_2(\hat \pi_{\theta\mid y}, \pi_{\theta\mid y})$ is equicontinuous in $y$.

Since $\mathcal Y$, also the support of $\pi_Y$, is compact, every open ball in $\mathcal Y$ has strictly positive $\pi_Y$-measure.
Now we can prove uniform convergence by contradiction.
Assume $\sup_{y\in\mathcal Y} W_2(\hat \pi_{\theta\mid y}, \pi_{\theta\mid y}) > \delta$ for some $\delta>0$, then there exists some $\bar y\in\mathcal Y$ such that $W_2(\hat \pi_{\theta\mid \bar y}, \pi_{\theta\mid \bar y}) > \delta$.
By equicontinuity, for any $y$ in the neighboring ball $B(\bar y,\epsilon)$ we have
\[
W_2(\hat \pi_{\theta\mid y}, \pi_{\theta\mid y}) >\delta - C\epsilon.
\]
If we pick $\epsilon = \delta/2C$, then we have $W_2(\hat \pi_{\theta\mid y}, \pi_{\theta\mid y}) >\delta/2$ within $y\in B(\bar y, \delta/2C)$.
But then
\[
\E_{\pi_Y} W_2^2 (\hat \pi_{\theta\mid y}, \pi_{\theta\mid y}) \geq \int_{B(\bar y, \delta/2C)} (\delta/2)^2 \pi_Y(\diff y),
\]
which is a positive constant unrelated to $N$.
This contradicts the $L^2$ convergence result.
Therefore we conclude that
\[
\sup_{y\in\mathcal Y}W_2 (\hat \pi_{\theta\mid y}, \pi_{\theta\mid y}) \rightarrow 0.
\]

\subsection{Proof of Corollary \ref{cor:credible.set.consistency} and Proposition \ref{prop:MK.conditional.rank}}

Note that
\[
W_2^2(\hat\pi_{\theta\mid y^*} , \pi_{\theta\mid y^*}) = \int \|\hat G(y^*,\theta_0) - G(y^*,\theta_0)\| p_0^\Theta(\diff \theta_0).
\]
By Theorem \ref{thm:MLP.consistency}, this quantity converges to zero uniformly over $y^*\in\mathcal Y$.
Since $\Theta$ is compact according to Assumption \ref{ass:farrell.1}, and both maps $G$ and $\hat G$ are Lipschitz on $\Theta$ as a result of Assumptions \ref{ass:flow.smoothness} and \ref{ass:spatial.lipschitz}, by the same technique as in the proof of Theorem \ref{thm:MLP.consistency}, we can derive the following uniform convergence guarantee that for any $y^* \in \mathcal Y$,
\begin{equation}\label{eq:G.uniform.conv}
\sup_{\theta_0\in\Theta} \| \hat G(y^*,\theta_0) - G(y^*,\theta_0)\| \rightarrow 0.
\end{equation}
Recall the definition of Bayesian credible sets in \eqref{eq:credible.set} and oracle sets in \eqref{eq:oracle.set}.
Consider any $\theta^*\in C_\tau(y^*)$.
Due to the compactness of $S^d(\tau)$, there exists $\theta^*_0 \in S^d(\tau)$ such that $\theta = G(y^*, \theta_0^*)$.
In fact, we can identify that $\theta^*_0$ as the Monge-Kantorovich conditional vector rank $R_{\theta\mid y^*}(\theta^*)$.
By transporting $\theta_0^*$ through the vector quantile map, we arrive at $\hat\theta^* := G(y^*, \theta_0^*)$.
From the definition \eqref{eq:credible.set}, $\hat\theta^*$ is in the credible set $\hat C_\tau(y^*)$.
Combining this argument with \eqref{eq:G.uniform.conv}, we conclude that for any $\theta^*\in C_\tau(y^*)$, there exists $\hat\theta^* \in \hat C_\tau(y^*)$ such that $\| \hat\theta^* - \theta^*\| \rightarrow 0$.
This implies
\[
\sup_{\theta \in C_\tau(y^*)}\inf_{\hat\theta\in \hat C_\tau(y^*)} \|\hat\theta - \theta\| \rightarrow 0.
\]
On the other direction, we can use the same proof technique and conclude that for any $\hat\theta^* \in \hat C_\tau(y^*)$, there exists $\theta^*\in C_\tau(y^*)$ such that $\| \hat\theta^* - \theta^*\| \rightarrow 0$, and thus
\[
\sup_{\hat\theta\in \hat C_\tau(y^*)}\inf_{\theta \in C_\tau(y^*)} \|\hat\theta - \theta\| \rightarrow 0.
\]
Due to the uniform convergence guarantee in \eqref{eq:G.uniform.conv}, the choice of $y^*\in\mathcal Y$ is arbitrary.
The choice of $\tau \in (0,1)$ is also arbitrary.
Therefore,
\[
\sup_{y^*\in\mathcal Y} d_H(\hat C_\tau(y^*), C_\tau(y^*)) \rightarrow 0.
\]
This finishes the proof of Corollary \ref{cor:credible.set.consistency}.

For the proof of Proposition \ref{prop:MK.conditional.rank}, observe that under the estimated posterior distribution recovered by $\hat G(y^*,\cdot)$, the event $\{\hat r_{\theta\mid y^*}(\theta) > 1-\alpha\}$ is equivalent to $\{\theta_0 \in S^d(1)\setminus S^d(1-\alpha)\}$.
Therefore,
\[
\hat\pi_{\theta\mid y^*}(\hat r_{\theta\mid y^*}(\theta) > 1-\alpha) = p_0^\Theta(S^d(1)) - p_0^\Theta (S^d(1-\alpha)) = \alpha.
\]
The uniform convergence in Theorem \ref{thm:MLP.consistency} implies weak convergence $\hat\pi_{\theta\mid y^*} \Rightarrow \pi_{\theta\mid y^*}$.
Recall that $\Theta$ is a compact domain, and the continuity of the maps are ensured by Assumptions \ref{ass:flow.smoothness} and \ref{ass:spatial.lipschitz}.
Denote
\[
\hat F(\tau) = \hat\pi_{\theta\mid y^*}(\hat r_{\theta\mid y^*}(\theta) \leq \tau), \quad F(\tau) = \pi_{\theta\mid y^*}(\hat r_{\theta\mid y^*}(\theta) \leq \tau).
\]
From the weak convergence and continuity, we get
\[
|\hat F(1-\alpha) - F(1-\alpha)| \rightarrow 0.
\]
This implies that $\pi_{\theta\mid y^*} (\hat r_{\theta\mid y^*}(\theta) > 1-\alpha) \rightarrow \alpha$, finishing the proof of Proposition \ref{prop:MK.conditional.rank}.

\section{ADDITIONAL EXPERIMENTS}\label{app:additional_exp}
To complement the evaluation in the main text, this section presents targeted experiments that probe two aspects of the proposed flow-matching posterior sampler: (1) monotonicity and recovery of the Bayesian credible set (Section \ref{app:monotonicity}) and (2) the effect of scaling for recovering the true posterior (Section \ref{sec:scaling}).

\subsection{Posterior for Neal's Funnel}
Here, we revisit the Neal's funnel (Section~\ref{sec:main-exp}) for posterior sampling. As described earlier, Neal's funnel is characterized by (\ref{eq:neals.funnel}).
Accordingly, by rearranging (\ref{eq:neals.funnel}), the posterior distribution is known up to normalizing constant:
\begin{equation}\label{eq:neals-funnel-posterior}
p(\nu \mid x) \propto \exp\left[ -\frac{1}{2} \left( \frac{\nu^2}{3^2} + e^{-\nu} x^2 + \frac{\nu}{2} \right) \right].
\end{equation}

\begin{figure}
    \centering
    \includegraphics[width=1\linewidth]{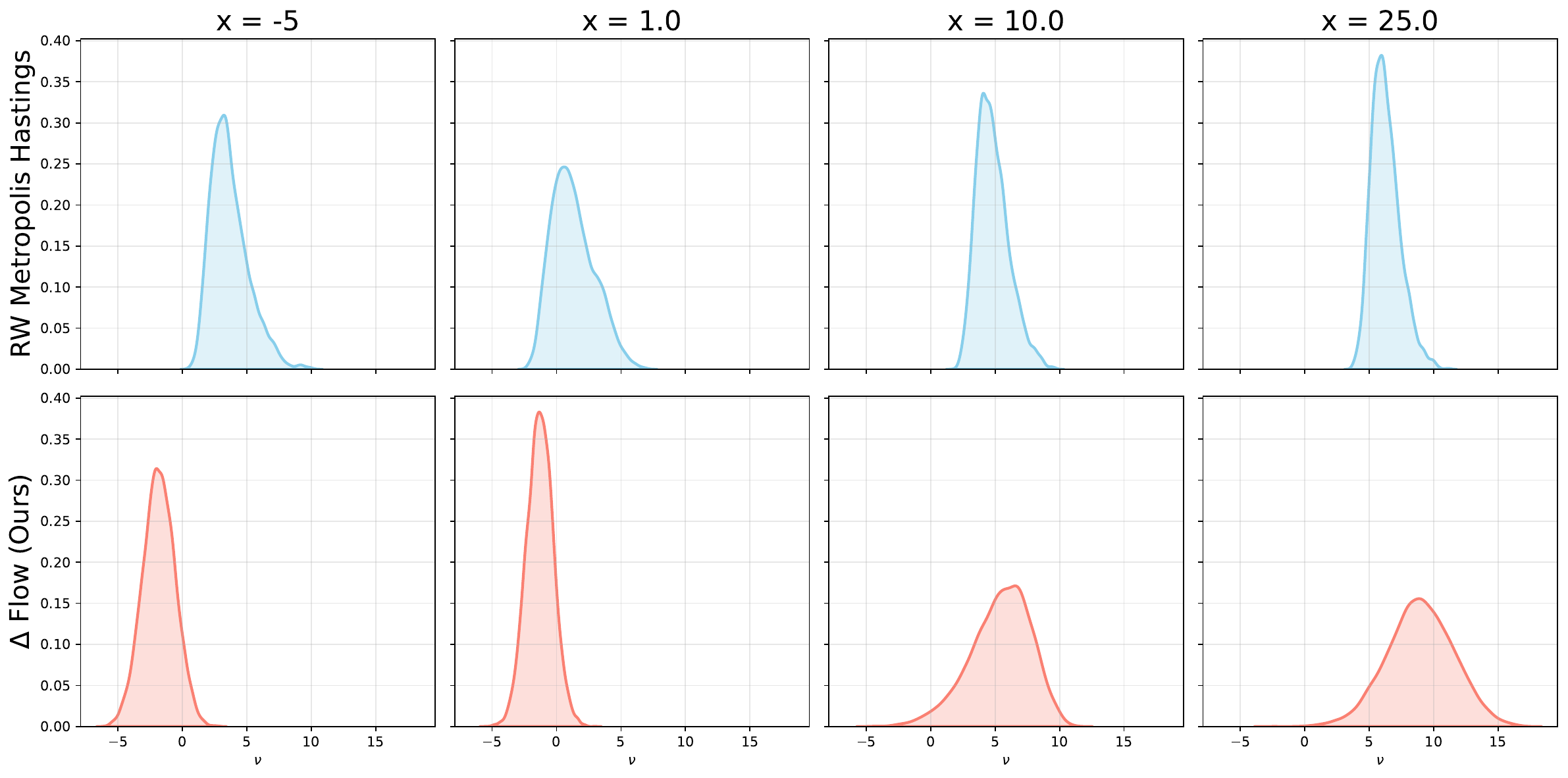}
    \caption{Posterior recovery of Neal's funnel (\ref{eq:neals.funnel}) at different observation $x = [-5, 1, 10, 25]$, compared with standard Metropolis-Hastings (MH) sampler (first row). Overall, the learned posterior by flow (second row) matches with MH estimations across the practical range of $x$.}
    \label{fig:neals_funnel_posterior}
\end{figure}

The posterior distribution can be well estimated by Metropolis-Hastings (MH) with symmetric proposal proposal
\begin{equation*}
    \nu^{*} \sim q(\cdot|\nu^{(t-1)}) = \mathcal{N}(\nu^{(t-1)}, \sigma^2),
\end{equation*}
with corresponding acceptance probability being
\begin{equation*}
    \alpha = \min \big\{ 1, \exp \big ( \ell(\nu^{*};x) - \ell(\nu^{(t-1)};x) \big ) \big\},
\end{equation*}
where $\ell$ is the log posterior (unnormalized) (\ref{eq:neals-funnel-posterior}). 
We evaluate the posterior distribution on different $x$ values. Here, the problem is favorable for MH as unnormalized posterior distribution is log concave in $\nu$ given $x$ (\ref{eq:neals-funnel-posterior}). Overall, the learned posterior by flow is a faithful surrogate across the practical range of $x$. For $x = 25$, where $\nu$ is likely to be from a large value and the data is scarce, flow model expands the tail slightly more compared to MH, see Figure~\ref{fig:neals_funnel_posterior}.

\subsection{Effect of Proper Scaling}\label{sec:scaling}

Recall Assumption \ref{ass:spatial.lipschitz}, which states that the approximate flow $\hat{v}_{t}(x)$ is differentiable with respect to $x$ and $t$, and for each $t$, there exists a constant $L_t$ such that $\hat{v}_{t}(x)$ is $L_{t}$-Lipschitz in $x$. In other word, for every fixed $t$, there exists $L_t \in \mathbb{R},
$ \begin{equation*}
    \|v_t(x) - v_t(x')\| \leq L_t \|x-x'\| \quad \forall x, x' \in \mathbb{R}^{n + p}.
\end{equation*}
The posterior consistency relies on Theorem \ref{thm:posterior.W2.bound}, which asserts that under Assumption \ref{ass:flow.smoothness} and \ref{ass:spatial.lipschitz}, the expected squared 2-Wasserstein distance between the estimated and true posterior is bounded by $2\epsilon_{N}^{2}Le^{2L}$, where $L=\int_{0}^{1}L_{t}dt$.
It shows that the sampling error exhibits an exponential growth with the Lipschitz constant $L_t$. 

For $x_t = [ y_t^\top, \theta_t^\top]^\top = [y_{1t},...,y_{nt}, \theta_{1t},...,\theta_{pt}]^\top$, we can denote the mean and covariance as $\mu_t = \mathbb{E} x_t$ and $\Sigma_t = \text{Cov}(x_t) = \mathbb{E}[ (x_t-\mu_t) (x_t- \mu_t)^\top]$. Because $x_t$ follows the deterministic flow $\frac{\diff}{\diff t}x_t = v_t(x_t)$, we could write the time derivative of covariance $\Sigma_t$ as 
\begin{equation}\label{eq:lipschitz.and.data.covaraince}
    \frac{\diff}{\diff t} \Sigma_t = \mathbb{E} \big [  (\nabla_x v_t)(x_t) \Sigma_t + \Sigma_t (\nabla_x v_t)^\top (x_t)\big  ].
\end{equation}
Since Assumption~\ref{ass:spatial.lipschitz} implies that the Lipschitz constant is a gradient in magnitude in that, for fixed $t$, 
\begin{equation}\label{eq:lipschitz.gradient}
    L_t = \sup_{x \neq x'}\frac{\|v_t(x) - v_t(x')\|}{\|x-x'\|} = \sup_x \|\nabla_x v_t(\cdot)\|_\text{op},
\end{equation}
and thus $\|\nabla_x v_t(x)\|_\text{op} \leq L_t$. 
Combining (\ref{eq:lipschitz.and.data.covaraince}) with (\ref{eq:lipschitz.gradient}), we obtain
\begin{align*}
     \frac{\diff}{\diff t} \Sigma_t  \preceq 2L_t \Sigma_t \quad \Leftrightarrow \quad \Sigma_t \preceq \exp \Big (2 \int_0^t L_s ds \Big ) \Sigma_0, \quad t \in [0,1]\
\end{align*}
where $\Sigma_0 = I_{n+p}$ when we start from independent source distribution.

Any entries of $\nabla_x v_t$ numerically being large pushes the Lipschitz constant $L_t$ up (\ref{eq:lipschitz.gradient}). For example, if $j$-th coordinate of $x$ has a tiny scale, e.g. $\sigma(x^{(j)}) = 10^{-3}$, while other coordinates are order 1. A change of $\pm 10^{-3}$ along $j$-axis is negligible in Euclidean distance, $\|x- x'\|$. However, if the network output varies by even a modest amount in that direction, the quotient can blow up 
\begin{equation*}
    \frac{|v_t^{(i)}(x)-v_t^{(i)}(x')|}{|x^{(j)} - x'^{(j)}|} \approx \frac{0.05}{10^{-3}} = 50,
\end{equation*}
meaning a single column of $\nabla_x v_t$ already contributes 50 to $\|\nabla_x v_t\|_\text{op}$. 
Intuitively, a skinny coordinate makes gradients look artificially big and inflates the Lipschitz constant $L_t$. 
Rescaling the thin coordinate to unit variance removes the magnifying effect and lowers Lipschitz constant over $t$.

In optimization perspective, since neural velocity fields are trained by stochastic optimizer, if one coordinate has exceptionally large gradient it would not properly updating across the coordinates. Standardizing $x_1$ --- or equivalently, reparameterizing $y$ and $\theta$ so that their empirical covariance is close to the identity ($\widehat{\Sigma}_1 \approx I$) --- restores balance and enables $\hat{v}_t$ to be trained effectively toward the target objective.

\begin{figure}
    \centering
    \includegraphics[width=1\linewidth]{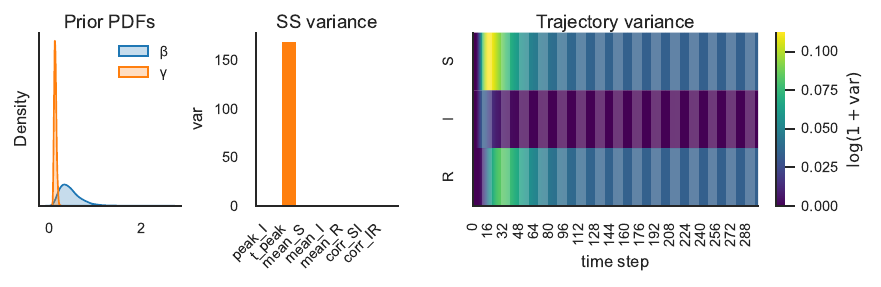}
    \caption{The prior for the infection rate $\beta$ is much more dispersed than the prior for the recovery rate $\gamma$: the sample variance of $\beta$ is 0.053 whereas $\gamma$ is only 0.0007 (ratio $\approx 79 : 1$).
    Among the seven handcrafted summary statistics, dispersion differs by two orders of magnitude: the variance of the epidemic peak size (peak I) is 166, while the growth-slope and correlation summaries are all below 0.05.
    Point-wise trajectory variances $\text{Var}\bigl(S_t,I_t,R_t\bigr)$ are shown on a $\log(1+\mathrm{var})$ scale.  The susceptible (S) and recovered (R) compartments start around 0.11 and decay steadily, whereas the infectious component (I) remains almost deterministic throughout, never exceeding 0.003 (below the first colour step).  
    Together, the three panels illustrate that the statistical scales relevant to learning the joint flow range from $10^{-3}$ (I-variance) to $10^{2}$ (peak I variance).}
    \label{fig:SIR-variance-range}
\end{figure}

\begin{figure}
    \centering
    \begin{subfigure}{0.45\textwidth}
    \centering
    \includegraphics[width=1\linewidth]{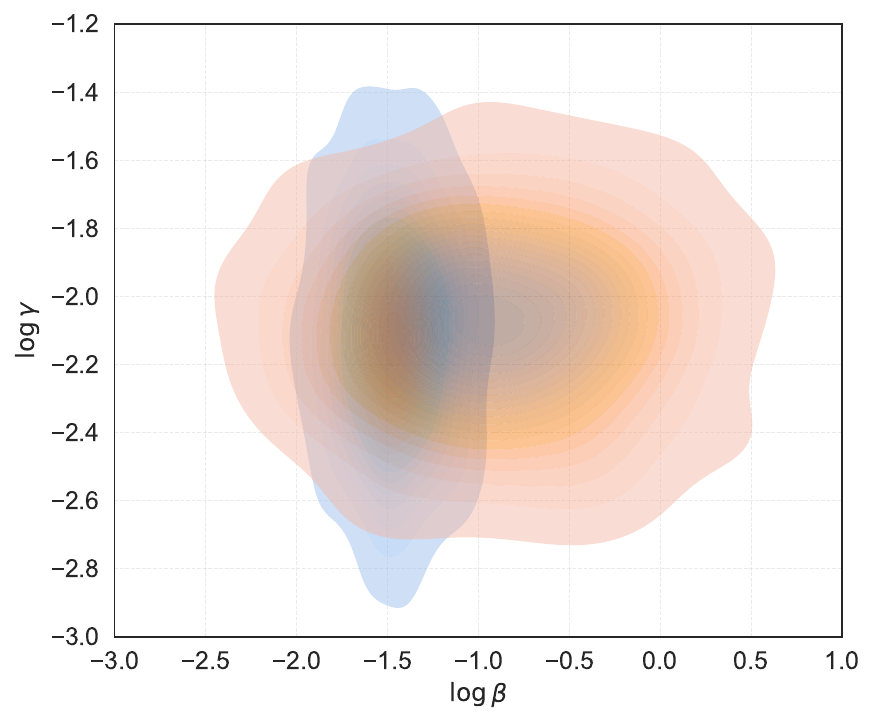} 
    \caption{Without Standardization}
    \end{subfigure}
    \begin{subfigure}{0.45\textwidth}
    \centering
    \includegraphics[width=1\linewidth]{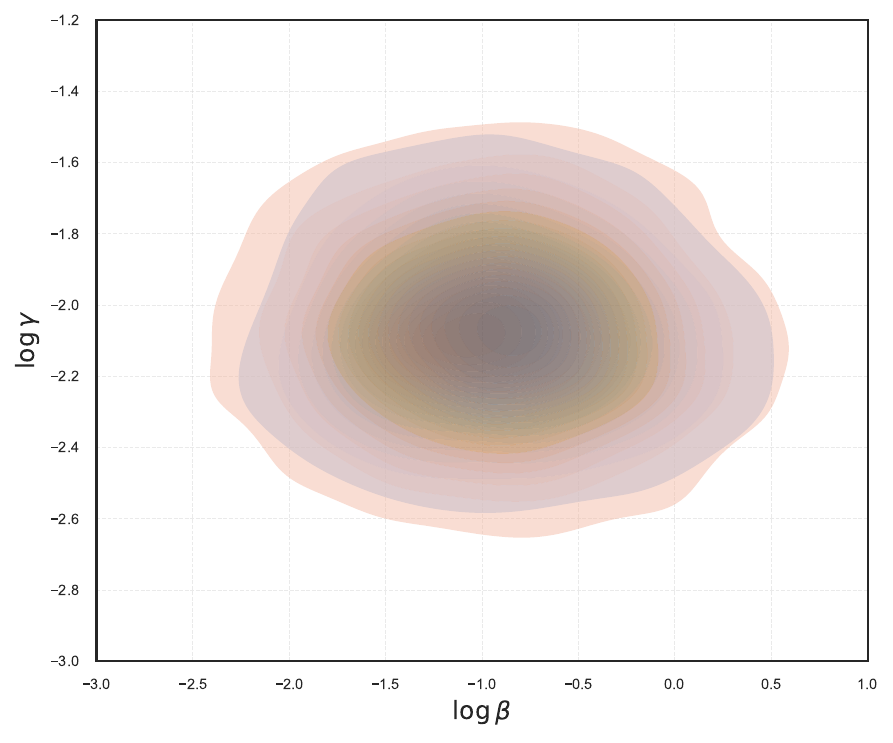}
    \caption{With Standardization}
    \end{subfigure}
    \caption{The figures show the estimated (blue) and true prior distribution (orange). Our method hinges on the correct estimation of joint distribution $(y, \theta) \sim L (y|\theta) \pi(\theta) $, and if the joint estimation is not correct, we cannot ensure the quality of posterior estimation. Note the effect of standardization on correctly estimating the prior distribution. Without normalization (a), the model even fail to learn the prior distribution for the parameters.}
    \label{fig:SIR-normalization-effect}
\end{figure}

Inspired from this theoretical observation (\ref{eq:lipschitz.gradient}), we show that appropriate rescaling on the joint space $\mathcal{Y}\times\Theta$ such that each dimension possesses similar variance is important in learning the joint distribution, by effectively aiding in controlling the Lipschitz constant through an SIR example, where the range of two spaces ($\mathcal{Y}$ and $\Theta$) are highly heterogeneous. 
See the differences in variance for the parameters and manual summary statistics space in Figure~\ref{fig:SIR-variance-range}. The effect of proper standardization is shown in Figure~\ref{fig:SIR-normalization-effect} in terms of an accuracy of joint recovery.

\subsection{Effect of Varying Priors on SIR Model}\label{app:varying_prior}

 \begin{figure}
    \centering
    \begin{subfigure}{0.6\textwidth}
    \includegraphics[width = \linewidth]{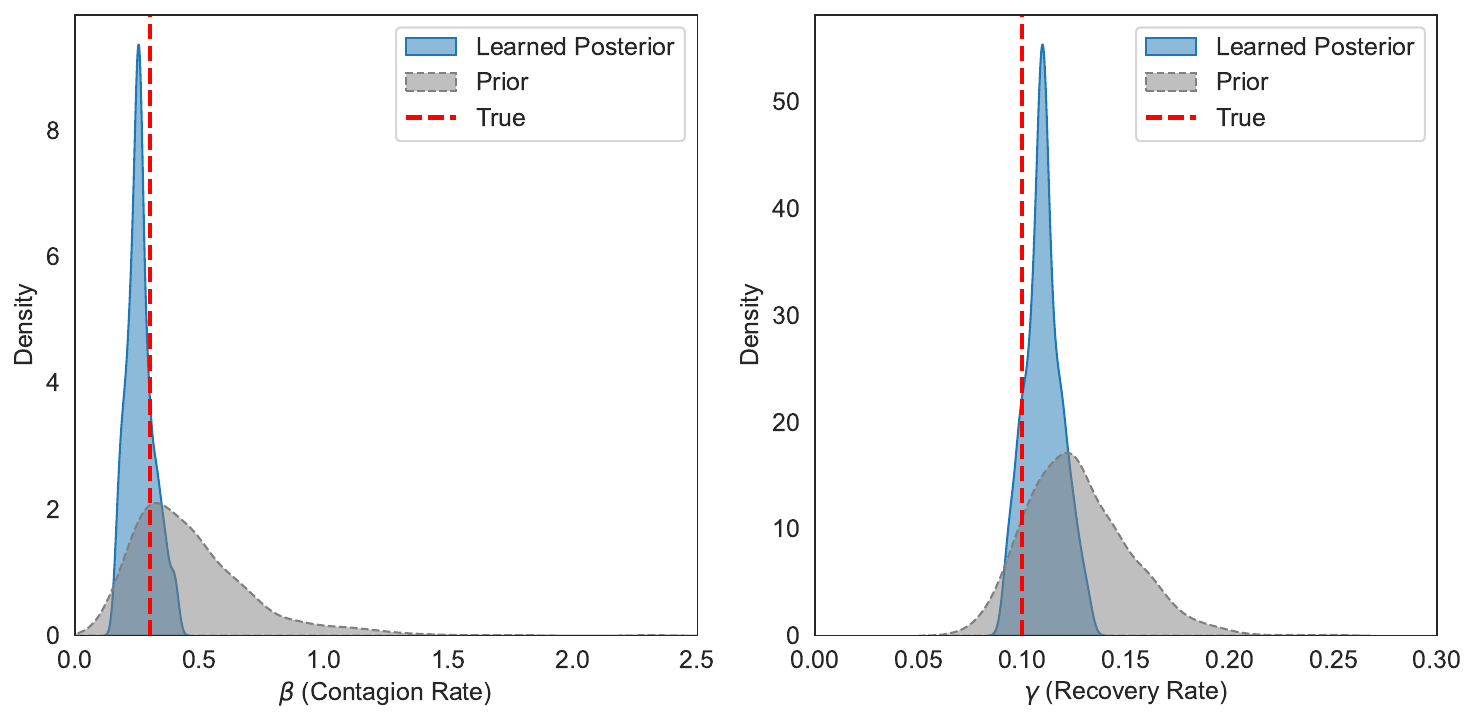}
    \caption{Marginal Posterior distribution}
    \label{fig:sir-contraction}
    \end{subfigure}
    \begin{subfigure}{0.33\textwidth}
    \includegraphics[width = \linewidth]{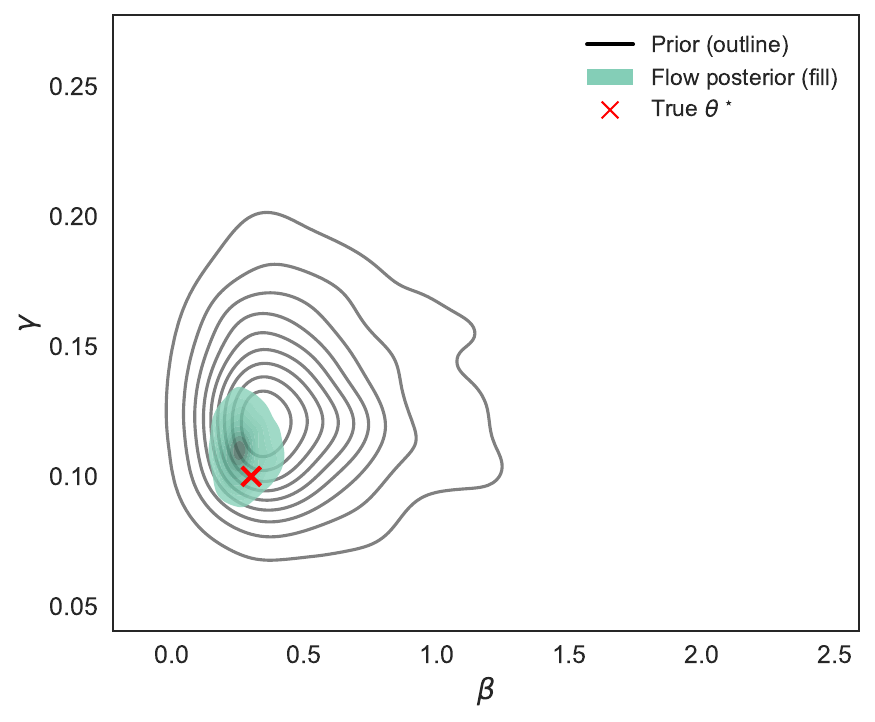}
    \caption{Joint Posterior estimation}
    \label{fig:sir-joint}
    \end{subfigure}
    \caption{ 
    The log-normal prior for both parameters $\beta$ and $\gamma$.
    The posterior (blue) for the parameters ($\beta, \gamma$) compared to their prior (grey) are given. The estimated posterior regions include the true parameter values (red dotted line or cross). }
    \label{fig:sir-model}
\end{figure}

\begin{figure}
    \centering
    \begin{subfigure}{0.6\textwidth}
    \includegraphics[width = \linewidth]{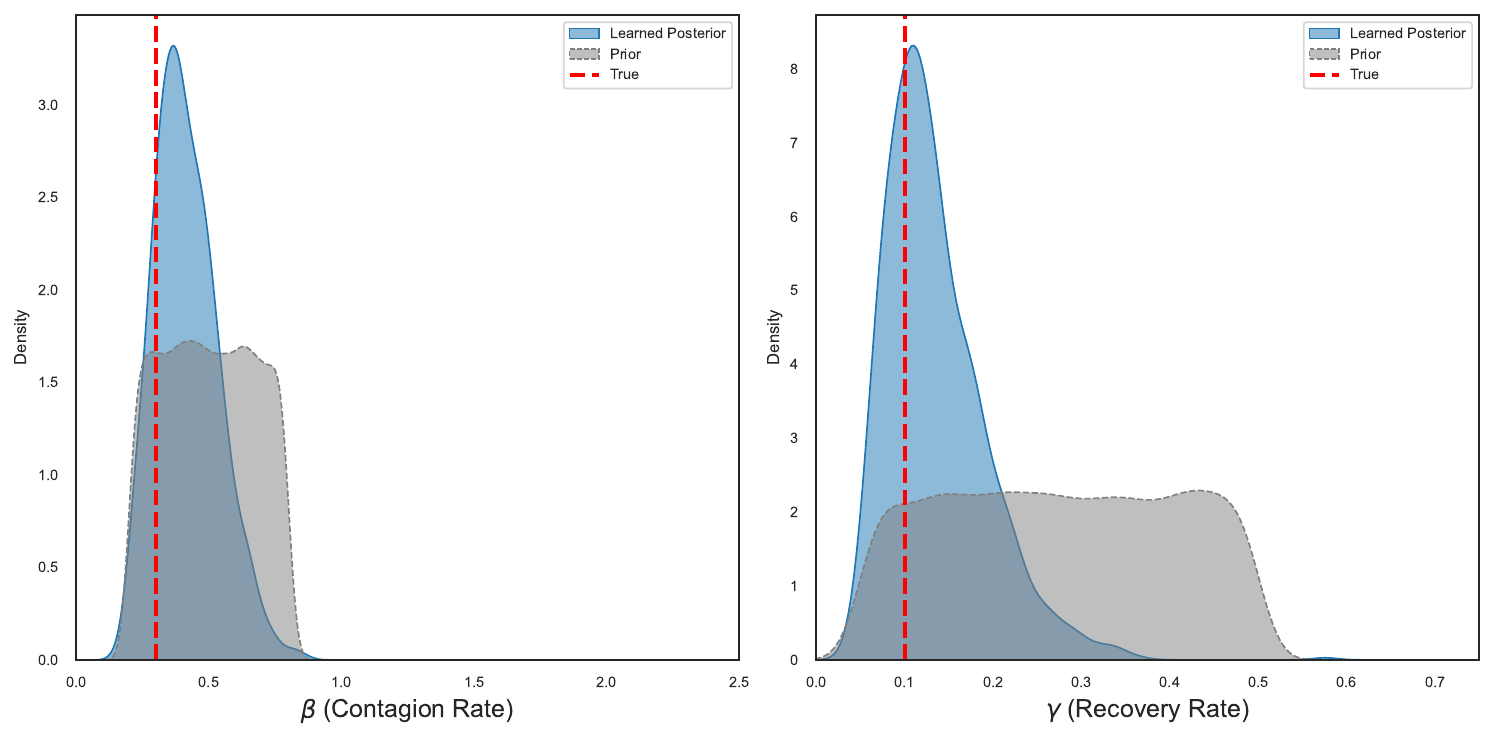}
    \caption{Marginal Posterior distribution with Uniform prior}
    \label{fig:sir-marginal-uniform}
    \end{subfigure}
    \begin{subfigure}{0.33\textwidth}
    \includegraphics[width = \linewidth]{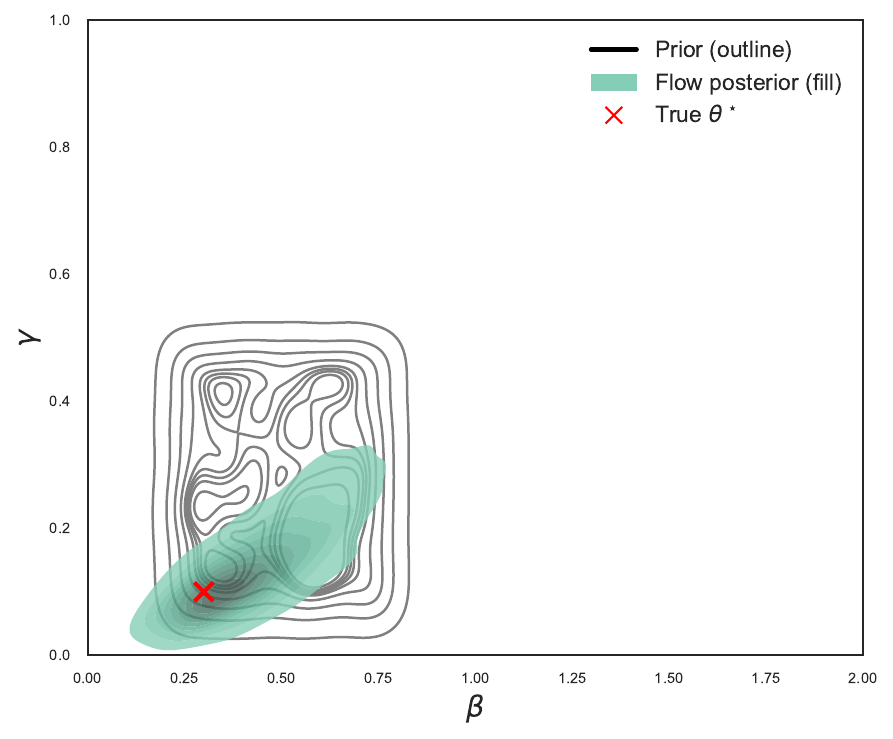}
    \caption{Joint Posterior estimation}
    \label{fig:sir-joint-uniform}
    \end{subfigure}
    \caption{ 
    Supplement to Figure~\ref{fig:sir-model}. The log-normal prior has been replaced with uniform prior. }
    \label{fig:sir-model-uniform}
\end{figure}

\begin{figure}
    \centering
    \begin{subfigure}{0.6\textwidth}
    \includegraphics[width = \linewidth]{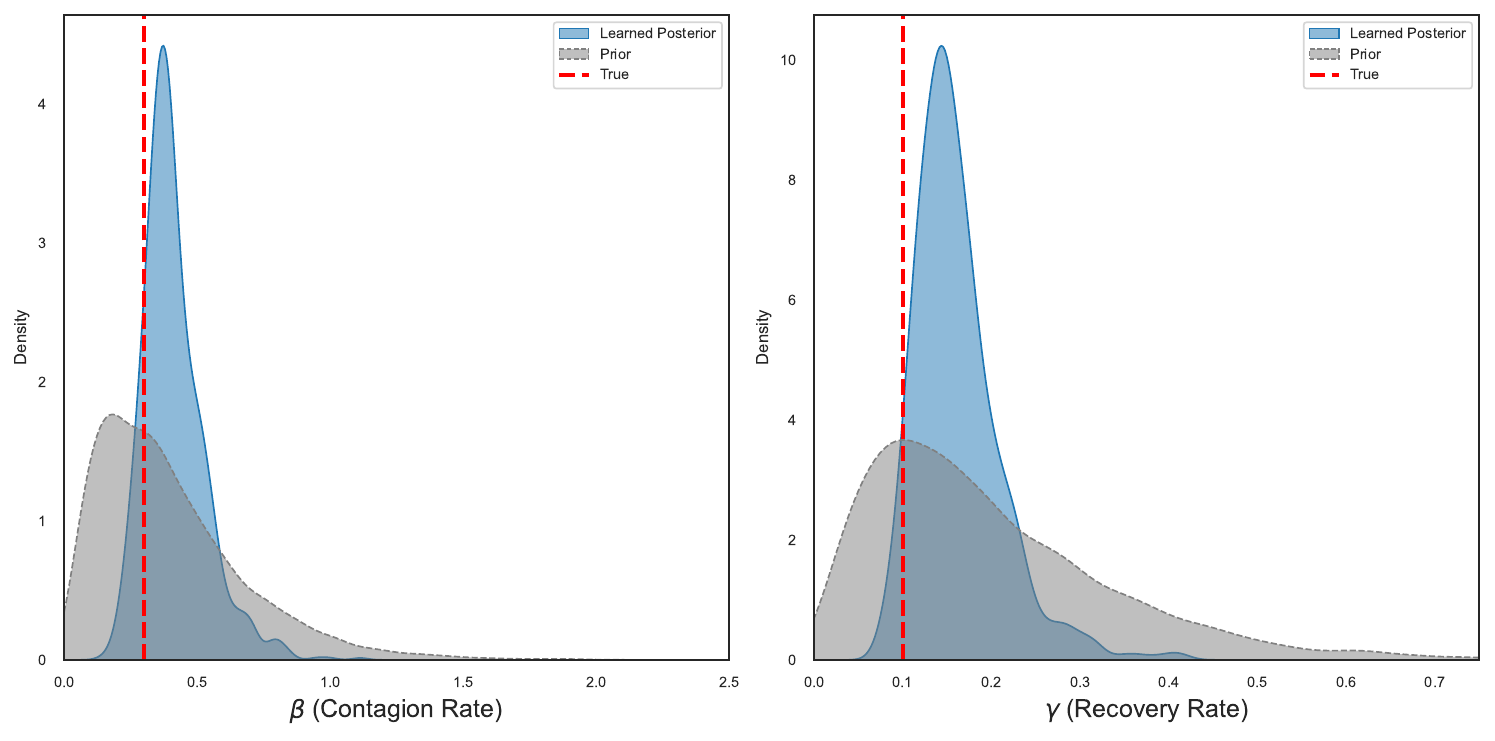}
    \caption{Marginal Posterior distribution  with Gamma prior}
    \label{fig:sir-marginal-gamma}
    \end{subfigure}
    \begin{subfigure}{0.33\textwidth}
    \includegraphics[width = \linewidth]{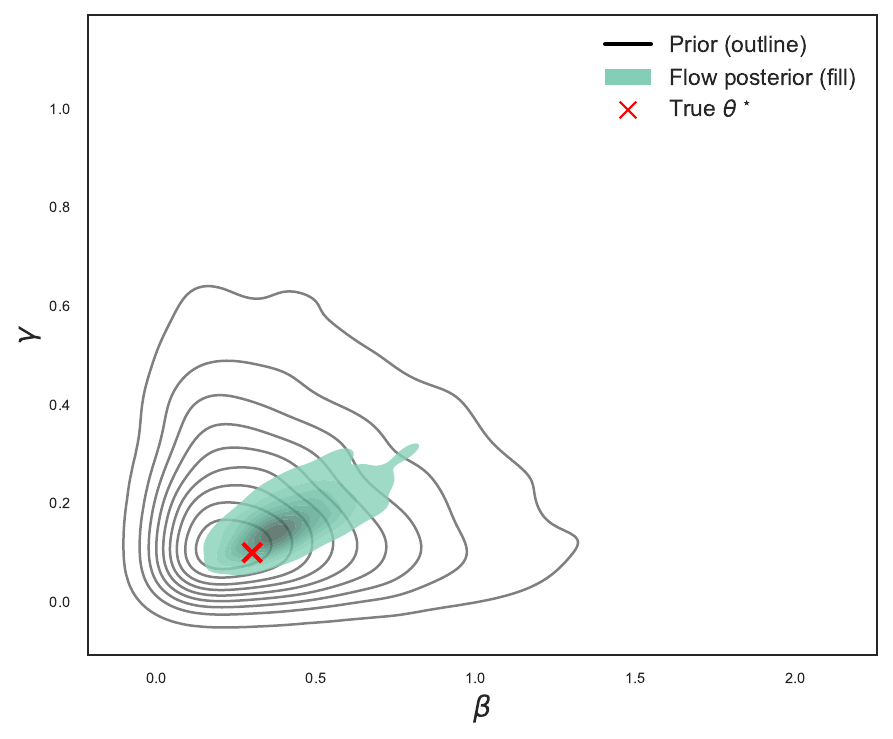}
    \caption{Joint Posterior estimation}
    \label{fig:sir-joint-gamma}
    \end{subfigure}
    \caption{ 
    Supplement to Figure~\ref{fig:sir-model}. The log-normal prior has been replaced with gamma prior. }
    \label{fig:sir-model-gamma}
\end{figure}

To examine the sensitivity of posterior inference to the choice of prior distribution, we conduct an additional experiment using the classical Susceptible-Infected-Recovered (SIR) epidemiological model. 
The SIR model describes the evolution of three compartments-susceptible, infected, and recovered, according to a nonlinear system of ordinary differential equations parameterized by a rate of contagion $\beta$ and a mean recovery rate $\gamma$. 
Specifically, the model is defined with a system of ODEs, 
\[
\frac{\diff}{\diff t}[S,I,R]
= \left[
-\frac{\beta S}{N} I ,
\frac{\beta S}{N}I - \gamma I ,
\gamma I \right],
\]
where $S$ is the number of individuals susceptible to be infected, $I$ is the number of individuals infected, $R$ is the number of individuals recovered from the disease, and $S+I+R$, the total population, is fixed.
For all experiments, we simulate trajectories over a fixed time horizon of 300 days using normalized initial conditions, with a fixed total population size.

We consider three distinct prior families over the parameters ($\beta, \gamma$), each commonly used in the epidemiological and simulation-based inference literature. 
The first is a log-normal prior (Figure~\ref{fig:sir-model}), where $\log \beta$ follows a normal distribution with mean $\log 0.4$ and standard deviation $0.5$, and $\log \gamma$ follows a normal distribution with mean $\log (1/8)$ and standard deviation $0.2$. 
The second is a uniform prior (Figure~\ref{fig:sir-model-uniform}), in which $\beta$ is sampled uniformly between $0.2$ and $0.8$, and $\gamma$ between $0.05$ and $0.5$, representing uninformative priors over a reasonable parameter range. 
The third is a gamma prior (Figure~\ref{fig:sir-model-gamma}), where $\beta$ is drawn from a Gamma distribution with shape $2.0$ and scale $0.2$ (mean $0.4$), and $\gamma$ from a Gamma distribution with shape $2.0$ and scale $0.1$ (mean $0.2$), capturing skewed distributions with heavier right tails.

For each prior, we simulate 5,000 trajectories of the SIR system by sampling parameter values from the prior and integrating the system forward in time. 
We then compute a fixed set of handcrafted summary statistics for each trajectory, including peak infection level, time to peak, average number of the susceptible, infected and recovered, and cross-correlations. 
These summary statistics serve as inputs to a flow matching model, which is trained to learn a mapping from summary statistics to posterior parameters. 
The flow model architecture and other training hyperparameters are held fixed across prior settings to isolate the impact of the prior alone.

Results of this experiment indicate that the learned posterior distributions are substantially influenced by the choice of prior, even when the ground-truth parameters remain constant. 
When trained under the log-normal prior, the posterior concentrates more tightly around the true values ($\beta = 0.3, \gamma = 0.1$), with moderate skewness reflecting the asymmetric uncertainty encoded by the prior. 
Under the uniform prior, the resulting posterior is more diffuse, particularly in directions orthogonal to the dominant likelihood ridge, due to the lack of structural regularization from the prior. 
The gamma prior induces posteriors with visibly heavier right tails, especially in $\gamma$, consistent with the prior's long-tailed density and its interaction with the forward dynamics.

\subsection{Monotonicity via Input Convex Neural Network (ICNN) \citep{amos2017inputconvexneuralnetworks}}\label{app:monotonicity}

Let $\psi: \mathbb{R}^n \times \mathbb{R}^p \times \mathbb{R} \to \mathbb{R}$ be a  Partial Input Convex Neural Network (PICNN) \citep{amos2017inputconvexneuralnetworks} that is convex in $\theta \in \Theta$ and is unrestricted for the rest $[y^\top, t] \in \mathcal{Y} \times [0,1]$. Define 
\begin{equation}\label{eq:g.monotone.formula}
    g_t(y, \theta) := \nabla_\theta \psi_t(y, \theta).
\end{equation}
Because the map $\theta \mapsto \psi_\cdot(\cdot, \theta)$ is $C^1$ and convex in $\theta$, its gradient is monotone with respect to $\theta$. 
Substituting the ordinary multilayer perceptron in $g_t$ with the gradient of convex function (\ref{eq:g.monotone.formula}) therefore imposes the required monotonicity with respect to $\theta$ by design, yet preserves full modeling flexibility in $y$ and $t$.

For training, we need to generate $x_0$ by sampling $y_0$ from $n$-dimensional spherical uniform that is $y_0 = r \times  \phi$, and sampling $\theta_0$ from a $d$-dimensional spherical uniform, that is $\theta_0 = \tau \times \xi$, where $\phi \sim \text{Unif}(\mathcal S^{n-1}(1))$, $\xi \sim \text{Unif}(\mathcal S^{d-1}(1))$, and $r,\tau \sim \text{Unif}[0,1]$.
This step is not needed, for example, when we use isotropic gaussian as our source distribution, where partial coordinate still follows the isotropic gaussian. 

For generating a credible set, as described in Section~\ref{sec:monotonicity}, given $y^*$, we can start from spherical uniform: we draw directions $\xi$ from spherical uniform distribution on $d$-dimensional unit sphere and fix the radius $\tau$, setting $\theta_0=\tau \times \xi$. 
Evolving $\theta_0$ under the joint velocity field $(f_t,g_t)$ under fixed $y^*$ transports the spherical shell $S^{d-1}(\tau)$ into the shell of the credible set $\partial\hat{C}_{\tau}(y^*) \subset\Theta$, which contains the parameter endpoints of all trajectories whose source norm equals $\tau$.
Monotonicity of $g_t$ guarantees radial ordering: if $\tau_1<\tau_2$ then every trajectory launched from the inner sphere remains inside the image of the outer sphere, hence $\hat{C}_{\tau_1}\subset\hat{C}_{\tau_2}$.
As a consequence the family $\{\hat{C}_{\tau}(y^*)\}_{0\leq \tau\leq 1}$ forms nested parameter regions whose posterior mass increases monotonically with $\tau$.
Choosing $\tau$ so that the spherical source contains probability $1-\alpha$ yields $\hat C_{\tau}$ a $(1-\alpha)$-Bayesian credible set.

\begin{figure}
    \centering
    \includegraphics[width=1\linewidth]{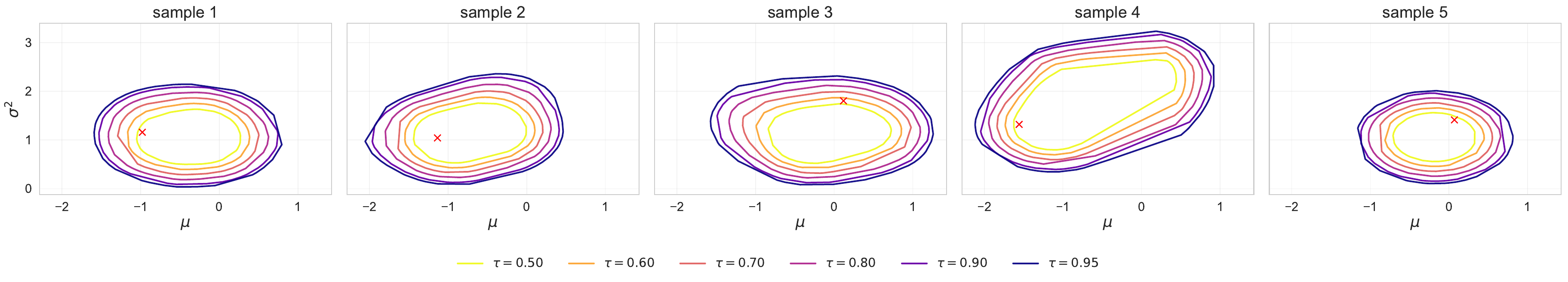}
    \caption{Bayesian credible set with increasing level of $\tau$ for Gaussian conjugate model described in Section~\ref{sec:main-exp}. Here, $n = 4$. Each column represents different realization of observed $X$. Observe that there is nested structure as we vary the $\tau$-level, and there is no crossing due to the monotonicity guaranteed by modeling through PICNN.}
    \label{fig:bayesian-credible-set}
\end{figure}

We first verify that the learned velocity indeed satisfies the monotonicity condition by checking that the inner product mentioned in Section \ref{sec:monotonicity} is non-negative on a dense grid of ($\theta,\theta'$) pairs.  
We then confirm the theoretical nesting property by inspecting credible sets obtained at successively larger radii: visualizations in parameter space show that sets corresponding to $\tau_{1}<\tau_{2}$ never intersect improperly, thereby validating both the monotonicity enforcement via the PICNN gradient and the credibility interpretation of the spherical-launch construction (Figure~\ref{fig:bayesian-credible-set}).

\begin{figure}
    \centering
    \includegraphics[width=0.6\linewidth]{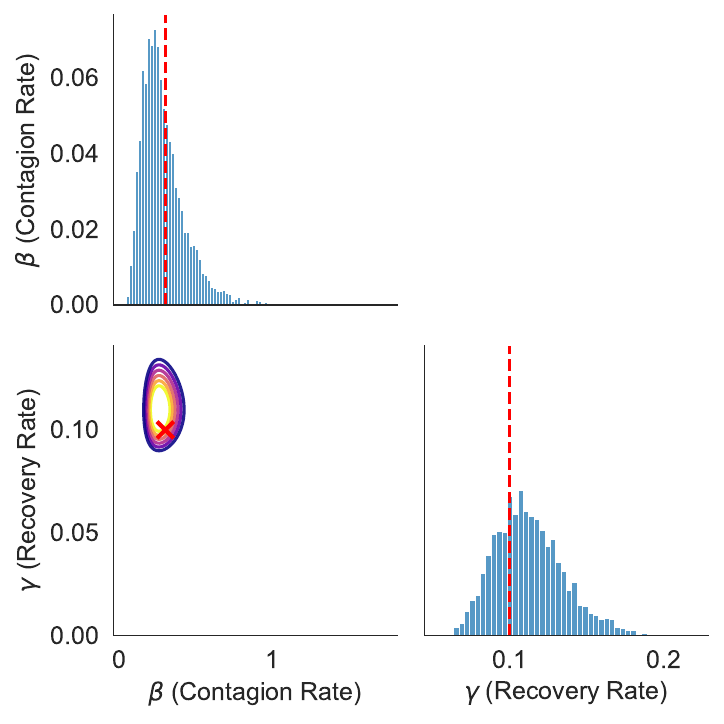}
    \caption{Estimated posterior distribution for $\beta$ (the contagion rate) and $\gamma$ (recovery rate). The corresponding $\tau = [0.5, 0.6, 0.7, 0.8, 0.9, 0.99]$ level credible sets are given as nested contours. The lightest contour corresponds to $0.5$-level credible set and the darkest contour corresponds to $0.99$-level credible set. The red vertical line on the histogram denotes the true parameter values ($(\beta,\gamma) = (0.3, 0.1)$). }
    \label{fig:SIR-credible-set}
\end{figure}

We revisit SIR example with additional monotonicity enforced on $g_t$. Figure~\ref{fig:SIR-credible-set} shows both marginal and joint posterior distribution covers the true underlying parameter values ($(\beta,\gamma) = (0.3, 0.1)$). Like in gaussian conjugate experiment (Figure~\ref{fig:bayesian-credible-set}), the estimated credible sets are non-crossing across different $\tau$ values and display nested structure. 

\subsection{Real Data Example: COVID-19 data analysis}\label{covid-analysis}

For our analysis we used publicly available daily COVID-19 surveillance data\footnote{https://www.kaggle.com/datasets/hadeelalqadi/uscovid-19data} from the Illinois Department of Public Health. The dataset spans March 10, 2020 to September 25, 2020 (200 days) and reports the date, total and new confirmed cases, total and new deaths, and the total number of individuals tested. 
From these quantities, we reconstructed the susceptible (S), infected (I), and recovered (R) compartments following the procedure in \cite{alqadi2022incorporating}.

The number of actively infected individuals on day t was defined as the rolling 14-day sum of newly confirmed infections, $I(t)=\sum_{k=t-13}^{t} N_k$. 
To estimate the susceptible population, we used both testing information and incubation dynamics. 
Assuming a 5-day incubation period, individuals who test positive on up to day $k+6$ are treated as susceptible on day $k$. Thus, $S(t)$ consists of those who test negative on day $t$ together with individuals who will test positive at least six days later, reflecting cases still in incubation. 
Because the dataset does not record recoveries, we approximated $R(t)$ by assuming that 95--99\% of confirmed infections eventually recover. Letting $D(t)$ denote cumulative deaths, the recovered compartment was estimated as
$R(t)= p\cdot \big( \sum_{k=t-15}^t N_k \big)- D(t),\; p\in[0.95,0.99]$. See the summary statistics of original data and reconstructed S,I,R value in Table~\ref{tab:covid-data-summary-statistics}.

\begin{table}[]
\centering
\renewcommand{\arraystretch}{1.1}
\resizebox{0.75\textwidth}{!}{
\begin{tabular}{@{}ccccccc@{}}
\toprule
\textbf{}     & \textbf{New cases ($N_t$)} & \textbf{Daily test} & \textbf{Death} & \textbf{S} & \textbf{I} & \textbf{R} \\ \midrule
\textbf{Mean} & 1,375                    & 25,286            & 4,950           & 32,336   & 18,372   & 111,496   \\
\textbf{SD}   & 819                    & 18,935            & 3,160           &20,294   & 10,147    & 77,738   \\
\textbf{Min}  & 0                          & 0                 & 0              & 14        & 0         & 0      \\
\textbf{Max}  & 5,594                      & 149,273             & 8,672         & 153,951     & 35,019     & 260,877    \\ \bottomrule
\end{tabular}
}
\caption{Descriptive Statistics of the Illinois COVID-19 Dataset (March 10, 2020-September 25, 2020)}
\label{tab:covid-data-summary-statistics}
\end{table}

In our analysis, we adopt log-normal priors on the transmission and removal rates,
$$\beta \sim \mathrm{LogNormal}(\log 0.4, 0.5^{2}),\qquad
\gamma \sim \mathrm{LogNormal}\big(\log (1/8), 0.2^{2}\big).$$
Given a parameter draw ($\beta,\gamma$), we simulate the SIR trajectories via numerical solution of the ODE system and train our triangular flow model on pairs ($\beta,\gamma$, \text{simulated data}). 
Once the model is trained, we generate posterior samples conditioned on the observed 200-day Illinois COVID-19 data.
To interpret the posterior we focus on epidemiologically meaningful functionals of $(\beta,\gamma)$: (1) the basic reproduction number $R_0=\beta/\gamma$, 
(2) the infectious period $D=1/\gamma$, 
(3) the early exponential growth rate $r=\beta-\gamma$, and (4) the corresponding doubling time $T_\mathrm{d}=\log 2 / r$. 

\paragraph{Results.} The recovery rate $\gamma$ implies an infectious period of approximately 14.6 days, consistent with published estimates of 10-20 days \citep{oelsner2024epidemiologic}.
The implied basic reproduction number $R_0 = \beta/\gamma \approx 3.5$ aligns with early-pandemic estimates for COVID-19 transmission in the United States, where $R_0$ values between 3 and 5 have been documented in the absence of interventions \citep{sanche2020high}. 
The MAP estimate of the early exponential growth rate $r = \beta - \gamma$ is 0.17 per day, corresponding to a doubling time is approximately $3.98$ days. 
This is consistent with empirical estimates of 2-4 day doubling times reported during the early, unconstrained phase of COVID-19 spread \citep{muniz2020doubling, pellis2021challenges}.
See Figure~\ref{fig:covid-19-posterior-stat} for the posterior distribution of each statistics and Table~\ref{tab:covid-map-estimate} for maximum a posteriori (MAP) estimate with 90\% marginal credible bound.
Although our estimates are not directly comparable to those in \cite{alqadi2022incorporating}, which rely on an extended SIR model with additional forcing terms, the overall magnitudes of the inferred parameters agree with the established epidemiological literature.


\begin{table}[t]
\centering
\begin{tabular}{lcccccc}
\toprule
 & $\beta$ & $\gamma$ & $R_0$ & $1/\gamma$ (days) & $r$ & $T_d$ (days) \\
\midrule
MAP 
& 0.24 
& 0.07 
& 3.54 
& 14.58 
& 0.17 
& 3.98 \\
LB 
& 0.12 
& 0.05 
& 1.53 
& 9.46 
& 0.04 
& 1.23 \\
UB
& 0.64 
& 0.11 
& 9.25 
& 18.75 
& 0.56 
& 14.26 \\
\bottomrule
\end{tabular}
\caption{Maximum A Posteriori (MAP) estimates of parameters and key statistics with 90\% marginal credible interval. Lower bound (LB) and Upper bound (UB) are given in the table.}
\label{tab:covid-map-estimate}
\end{table}

\subsection{Experiment Details}\label{sec:experiment-details}

Unless otherwise noted, every experiment is driven by the same \emph{source distribution} -- an isotropic Gaussian $\mathcal N(0,I_{n+d})$ in the joint space $\mathcal Y\times\Theta$.  
All code and exact configurations are available in our \href{https://github.com/sowonjeong/flow-posterior}{GitHub repository}.

\paragraph{Neal's Funnel}
Both the data marginal velocity $f_t$ and the conditional velocity $g_t$ are implemented as four-layer fully connected networks with hidden width $64$ and \textsc{ELU} activations.  
We train for $20,000$ iterations using Adam ($\text{lr}=0.001$) \citep{kingma2014adam}.  

\paragraph{Gaussian Conjugate}
The flow architecture mirrors that of the funnel experiment but with hidden width starting from  $64$ and increasing proportionally to the size of $n$; the lower dimensionality allows a lighter network without loss of accuracy.  

\paragraph{SIR}
Denote the state trajectory by $y=(S_t,I_t,R_t)_{t=1}^{T}$ and let $\theta=(\beta,\gamma)$.
The data marginal velocity $f_t$ is a four-layer MLP with hidden width $64$, while $g_t$ is the gradient of a two-block
\emph{spectrally normalized} ResNet,

where each residual block has width $128$ and \textsc{SiLU} activations.
Input preprocessing follows the pipeline in Section~\ref{sec:main-exp} and Section~\ref{sec:scaling}:
(i) the raw time series is collapsed to the seven-dimensional summary vector $\bigl[\text{peak}_I,\;t_{\text{peak}},
          \;\overline S,\;\overline I,\;\overline R,
          \;\operatorname{corr}(S,I),\;\operatorname{corr}(I,R)\bigr];$
(ii) every summary is standardized to zero mean and unit variance over the training batch; (iii) the parameter is log–transformed
$(\log\beta,\log\gamma)$ and then standardized.
Training uses AdamW \citep{loshchilov2019decoupledweightdecayregularization} with weight-decay $10^{-4}$, batch size $256$,
learning-rate $2\times10^{-4}$.

\paragraph{Coverage experiment}
In this experiment we assess the calibration of credible sets produced by our flow model under the Gaussian conjugate setup (see Section~\ref{sec:main-exp} for its exact setup). 
For each synthetic dataset $y$, we construct a candidate $\tau$-level credible set $C_\tau(y)$ by sampling $n_{\text{set}}=2000$ noise vectors uniformly within an inner $\tau$-ball, pushing them through the learned flow conditioned on the data, $y$, and taking the convex hull of the resulting $(\mu,\sigma)$-samples. 
This yields nested sets across $\tau\in\{0.5,0.6,0.7,0.8,0.9,0.95\}$ as in Figure~\ref{fig:bayesian-credible-set}. 
To evaluate whether $C_\tau(y)$ captures the intended posterior mass, we independently draw $M=2000$ posterior samples from the flow (conditioned on the same observation $y$) and compute the proportion that lies inside each convex hull.

We repeat this procedure for $100$ independent draws of $y$. The resulting coverage proportions are aggregated across repetitions and visualized as boxplots in Figure \ref{fig:credible_main}, alongside the ideal diagonal $y=\tau$. 
The figure shows that the empirical mass closely tracks the nominal level $\tau$.

\paragraph{Details for Table~\ref{tab:time_comparison} 
and Table~\ref{tab:posterior_benchmarks} 
}
For each task, we use the package \texttt{sbibm} \citep{lueckmann2021benchmarking}. 
We draw 10 ground-truth parameter and observation pairs from the prior and simulator, build 10k reference posterior samples per observation, and evaluate algorithms across simulation budgets from 1k to 100k. 
For rejection ABC and Sequential Monte-Carlo ABC (SMC-ABC) \citep{beaumont2009adaptive}, we also use the \texttt{sbibm} implementation.
Performance is reported primarily via classifier two-sample tests (C2ST) using an MLP with two hidden layers (width = 10$\times$data dimension) and five-fold cross-validation; runtimes are also recorded (Table~\ref{tab:time_comparison}).

For guided flow, we follow \cite{zheng2023guided} method with  fixed the guidance strength $\omega = 1.5$.

We adopt the following five benchmark models for evaluation, all of which has analytical solution for posterior computation we could compare against.

\begin{enumerate}
    \item \textbf{Gaussian Linear}
    \begin{itemize}
        \item Prior: $\mathbf{\theta} \sim \mathcal{N}(0, 0.1 \odot I)$, $\mathbf{\theta}\in \mathbb{R}^{10}$
        \item Likelihood: $\mathbf{x}\mid \mathbf{\theta} \sim \mathcal{N} (\mathbf{m}_\theta = \mathbf{\theta}, \mathbf{S} = 0.1 \odot I)$, $\mathbf{x} \in \mathbb{R}^{10}$
    \end{itemize}
    \item \textbf{Simple Likelihood and Complex Posterior (SLCP)}
    \begin{itemize}
        \item Prior: $\mathbf{\theta} \sim \text{Unif}(-3, 3)$, $\mathbf{\theta} \in \mathbb{R}^5$
        \item Likelihood: $\mathbf{x} \mid \mathbf{\theta} = (\mathbf{x}_1, \mathbf{x}_2, \mathbf{x}_3, \mathbf{x}_4)$, $\mathbf{x}_i \sim \mathcal{N}(\mathbf{m}_\theta, \mathbf{S}_\theta)$, where $\mathbf{m}_\theta = \begin{bmatrix}
            \theta_1 \\ \theta_2
        \end{bmatrix}$, $\mathbf{S}_\theta = \begin{bmatrix}
            s_1^2 & \rho s_1 s_2 \\ \rho s_1 s_2 & s_2^2
        \end{bmatrix}$, $s_1 =\theta_3^2$, $s_2 = \theta_4^2$, and $\rho = \tanh(\theta_5)$. $\mathbf{x} \in \mathbb{R}^8$
    \end{itemize}
    \item \textbf{Gaussian Mixture}
        \begin{itemize}
    \item Prior: $\mathbf{\theta} \sim \text{Unif}(-10, 10)$, $\mathbf{\theta} \in \mathbb{R}^2$.
    \item Likelihood: $\mathbf{x} \mid \mathbf{\theta} \sim 0.5 \mathcal{N}(\mathbf{x} \mid \mathbf{m}_\theta = \mathbf{\theta}, \mathbf{S} = \mathbf{I}) + 0.5\mathcal{N}(\mathbf{x} \mid \mathbf{m}_\theta = \mathbf{\theta}, \mathbf{S} = 0.01 \cdot \mathbf{I})$, $\mathbf{x} \in \mathcal{R}^2$.
    \end{itemize}
    \item \textbf{Bernoulli Generalized Linear Model (GLM)}
        \begin{itemize}
        \item Prior: $\beta \sim \mathcal{N}(0,2)$, $\mathbf{f} \sim \mathcal{N}\big( 0, (\mathbf{F}^\top \mathbf{F})^{-1} \big) $, $\mathbf{F}_{i,i-2} = 1$, $\mathbf{F}_{i,i-1} = -2$, $\mathbf{F}_{i,i} = 1+ \sqrt{\frac{i-1}{9}}$, $\mathbf{F}_{i,j} = 0$ otherwise, $1 \leq i,j \leq 9$.
        \item Likelihood: $\mathbf{x}\mid\mathbf{\theta} = (\mathbf{x}_1, \dots, \mathbf{x}_{10})$,
        $\mathbf{x}_1 = \Sigma_{i}^T z_i$, $\mathbf{x}_{2:10} = \frac{1}{x_1}\mathbf{V}\mathbf{z}$
        ${z}_i \sim \text{Bern}\big(\eta(\mathbf{v}_i^\top \mathbf{f} + \beta) \big) $, $\eta(\cdot) = \exp(\cdot)/(1+\exp(\cdot))$. 
    \end{itemize}
    \item \textbf{Two Moons}
    \begin{itemize}
        \item Prior: $\mathbf{\theta} \sim \mathcal{U}(-1,1)$, $\mathbf{\theta} \in \mathbb{R}^2$
        \item Likelihood: $\mathbf{x} \mid \mathbf{\theta} = \begin{bmatrix}
            r \cos (\alpha) + 0.25\\ r \sin (\alpha) 
        \end{bmatrix}$ + 
        $\begin{bmatrix}
            - | \theta_1 + \theta_2| / \sqrt{2}\\
            (-\theta_1 + \theta_2) / \sqrt{2}
        \end{bmatrix}$,
        where $\alpha \sim \mathcal{U}(-\pi/2, \pi/2)$, $r \sim \mathcal{N}(0.1, 0.01^2)$, $\mathbf{x} \in \mathbb{R}^2$
    \end{itemize}
\end{enumerate}

\paragraph{Computation}
The experiments in this study were conducted using a combination of personal and institutional computational resources. 
Preliminary analyses and prototyping were performed on a MacBook Pro with an Intel Core i7 processor and 16GB of RAM. 

For larger-scale experiments, we used high-performance computing resources provided by the institution's research cluster, which includes access to multi-core CPUs with 128GB of RAM.
We did not use any GPU for the experiments.
While execution time varied by dataset and task, typical runs for clustering and evaluation completed within a few hours.

\end{document}